\documentclass[11pt]{article}

\usepackage[final]{acl}
\usepackage{tcolorbox} 
\tcbuselibrary{breakable}
\usepackage{pgfplots}
\pgfplotsset{compat=1.18}
\usepackage{times}
\usepackage{latexsym}
\usepackage{booktabs}
\usepackage{mdframed}
\usepackage{listings}
\usepackage[T1]{fontenc}
\usepackage{tikz}
\usetikzlibrary{shapes.geometric, arrows.meta, positioning, calc}
\usepackage[utf8]{inputenc}
\usepackage{tcolorbox}

\usepackage{microtype}

\usepackage{inconsolata}

\usepackage{graphicx}
\usepackage{multirow}

\usepackage{xcolor}
\definecolor{seedingblue}{HTML}{E8F0FA}
\definecolor{darkgreen}{RGB}{0,100,0}
\definecolor{filteringorange}{HTML}{FE6100}
\definecolor{mtyellow}{HTML}{FFB000}

\usepackage{listings}

\title{FALCON: A Model and Dataset Agnostic Framework for Synthetic Data Generation for NL2SQL Pairs}

\author{
  Darian Lee$^*$, Shannon Rumsey$^*$, Jack St. Clair$^*$, Xinyi Tang$^*$ \\
  University of California, Santa Cruz \\
  1156 High Street, Santa Cruz, CA 95064 \\
  \texttt{\{daeilee, sarumsey, jrstclai, xtang21\}@ucsc.edu}
  \AND
  Aditya Bansal,
  Yuanming  Shi\footnotemark[2] \\
  Adobe \\
  \texttt{\{jeremyshi, adibansal\}@adobe.com} \\
  \\
}

\begin{document}
\maketitle
\renewcommand{\thefootnote}{\fnsymbol{footnote}}
\footnotetext[1]{Equal contribution.}
\footnotetext[2]{Work performed while at Adobe.}
\begin{abstract}
Relational databases are among the most widely deployed forms of structured knowledge, and natural language access to them requires grounding language onto schema entities and relations while handling the ambiguity inherent in how people phrase requests. Existing synthetic NL-to-SQL data generation methods largely ignore this ambiguity and produce oversimplified queries that fail to prepare models for the complexity of real-world structured knowledge access. We present FALCON, a framework that generates realistic, ambiguity-aware NL-to-SQL data matching the complexity of challenging real-world benchmarks, at low cost using compact open models. Our approach combines reserved-word SQL seeding and persona-based prompting to generate structurally complex queries, while alignment-based filtering preserves difficulty by distinguishing genuinely incorrect examples from complex but valid queries. Human evaluation confirms consistent high quality across model sizes, and our generated data exceeds existing benchmarks in both SQL complexity and natural language richness. Difficulty-stratified analysis shows models trained on FALCON data increasingly outperform baseline-trained models as query complexity increases, validating our pipeline's success in generating challenging training data. When combined with a small proportion of existing benchmark data, mixed training recovers performance on simpler queries while preserving these advantages on complex ones. The model- and database-agnostic design enables organizations to generate high-complexity NL-to-SQL training data locally without external APIs.

\end{abstract}

\section{Introduction}

As large language model (LLM) capabilities advance, natural language interfaces for databases are becoming increasingly accessible to non-SQL experts, making NL-to-SQL parsing an increasingly consequential knowledge-grounding task \cite{yang-etal-2024-synthesizing, dai-etal-2025-parsql, yu-etal-2018-spider, li2023bird, gao2024dailsql, first_nl_paper}. This requires resolving natural language references onto specific schema entities and relations, much as entity linking and relation extraction do for other forms of knowledge base construction. In practice, this grounding is complicated by real-world database users, who employ diverse linguistic styles, pose ambiguous requests, and work with complex SQL operations that existing benchmarks underrepresent \cite{omni, ambrosia, li2023bird, Floratou2024NL2SQLIA}.

While challenging datasets such as BIRD \cite{li2023bird} have driven progress toward more realistic benchmarks, they remain human-authored, limiting scalability and adaptability to new domains. Existing synthetic data generation approaches face a critical trade-off: they either rely on expensive frontier model APIs, making them cost-prohibitive for many organizations, or produce oversimplified queries that fail to capture production-level complexity \cite{lupidi2025source2synth, guo-etal-2025-sqlforge}.

We introduce \textbf{FALCON}, a \textbf{F}ramework for synthetic \textbf{A}mbiguity-aware \textbf{L}inguistically-diverse \textbf{C}ost-efficient \textbf{O}ptimized \textbf{N}L-to-SQL data generation. FALCON addresses three core challenges: (1) generating BIRD benchmark-level query complexity, (2) capturing realistic linguistic diversity including ambiguity, and (3) achieving low cost using only compact open models (e.g. Qwen-2.5-3B). The framework is designed to complement existing datasets by targeting the complex end of the query distribution where existing resources fall short, and is model- and database-agnostic, allowing organizations to generate domain-specific training data for their internal schemas without sending sensitive information to external APIs.

\section{Related Work}

\subsection{The Academic-Industry Gap in NL-to-SQL} 

Early NL-to-SQL datasets established foundational benchmarks but with significant limitations for real-world deployment. WikiSQL \citep{zhong2017seq2sqlgeneratingstructuredqueries, yavuz-etal-2018-takes} focused on single-relation queries, while Spider \citep{yu-etal-2018-spider} and BIRD \citep{li2023bird} introduced cross-domain complexity and professional-domain examples. However, recent work has identified critical gaps between benchmark performance and production viability, including ambiguous queries from informal language use and underspecification, and database complexity at enterprise scale \citep{Floratou2024NL2SQLIA}. 

\subsection{The Cost Barrier to Industry Deployment}

Beyond benchmark-to-production gaps, cost efficiency and privacy present fundamental barriers to enterprise adoption. While frontier LLMs demonstrate strong performance, their API costs make continuous deployment prohibitive \citep{hoang-etal-2025-distill}. Fine-tuning smaller models offers significant cost and privacy advantages \citep{10.14778/3750601.3750696, omni} and can outperform prompting \citep{dawn}, but requires substantial high-quality training data. Organizations need production-ready systems handling complex queries and diverse language \citep{dawn}, yet generating training data via expensive frontier models or human annotators undermines local deployment benefits.

\subsection{Synthetic Data Generation for NL-to-SQL}
Researchers have explored automated synthesis to overcome scalability limitations. Source2Synth \citep{lupidi2025source2synth} combines WikiSQL tables with LLM-generated facts using execution-based verification, but requires fine-tuning multi-billion parameter models for filtering and focuses only on single-table queries. OmniSQL \cite{omni} expands to more difficult queries and diverse linguistic styles, yet lacks ambiguity handling and requires larger model ensembles. SQLForge \citep{guo-etal-2025-sqlforge} introduces SQL template parsing and iterative domain exploration but relies on expensive closed-source models (GPT-4) with declining execution rates on complex queries.

\subsection{Handling Query Ambiguity}
Addressing ambiguous queries is critical for production reliability \cite{Floratou2024NL2SQLIA}, yet most synthetic datasets generate only unambiguous pairs. Know What I Don't Know \citep{wang-etal-2023-know} categorizes unanswerable questions but focuses strictly on schema-level ambiguity, excluding broader linguistic ambiguities. Ambrosia \citep{ambrosia} identifies database-independent ambiguities (scope, attachment, vagueness) that reflect real user behavior but remain underrepresented in training data \cite{Floratou2024NL2SQLIA}.

\paragraph{Implications} While these existing methods address specific facets of NL-to-SQL generation, there remains a need for a unified, cost-efficient framework that handles complexity diversity and ambiguity; we address these gaps with the FALCON pipeline described in the following section.

\section{Pipeline}
\begin{figure*}[t]
  \includegraphics[width=\textwidth]{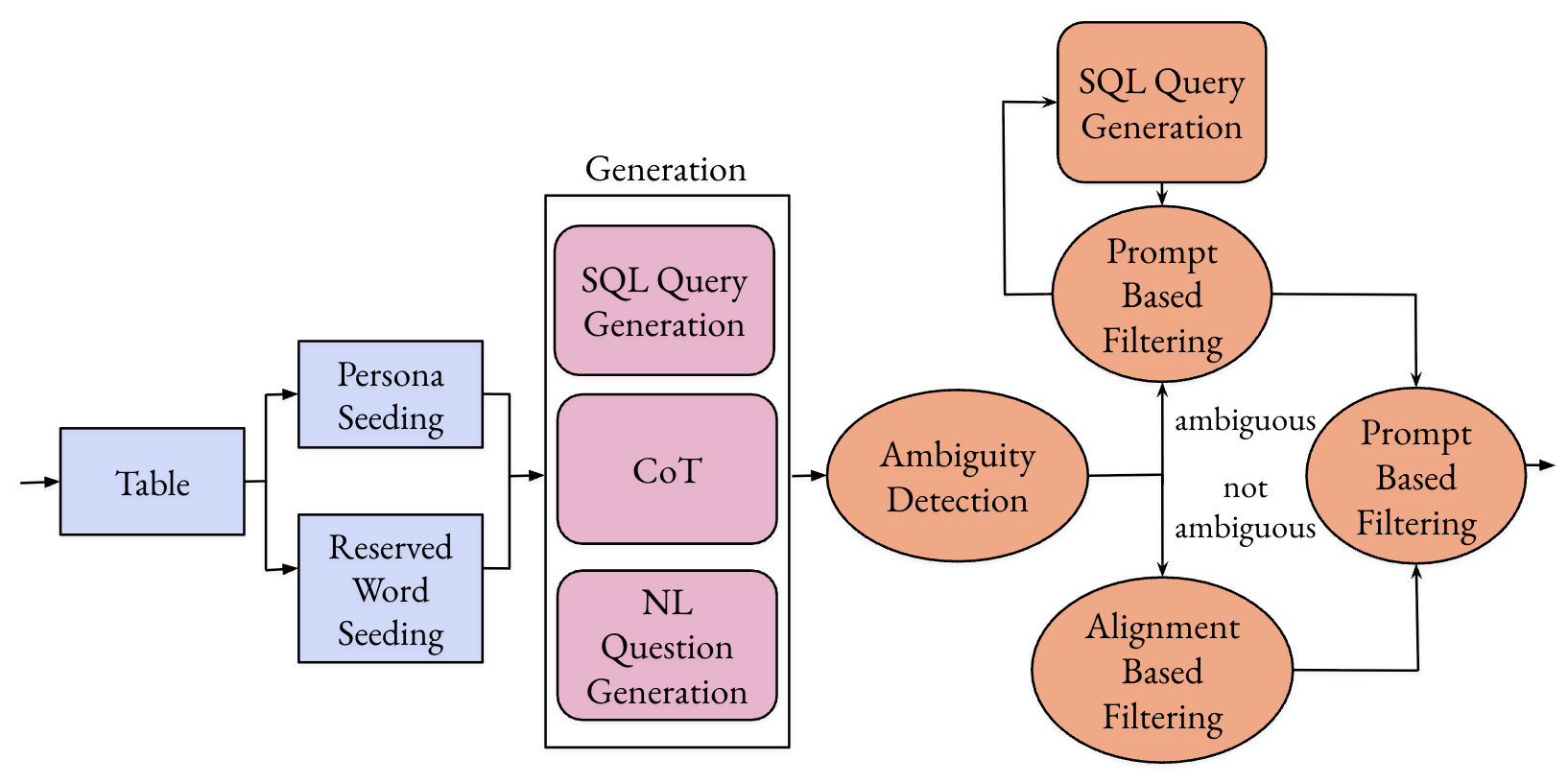}
  \caption{ Simplified architecture of the FALCON pipeline. It integrates systematic seeding, SQL and NL generation with CoT, and a multistage filtering process (Ambiguity, Reward, and Prompt-based), to produce high-quality NL-to-SQL pairs. To see a more detailed diagram of our pipeline, see Figure~\ref{fig:pipeline_flowchart}}
  \label{fig:pipeline}
\end{figure*}

Our pipeline follows a multi-stage generation and filtering process: (1) SQL generation with reserved word and persona seeding, followed by executability filtering; (2) NL generation with LLM-as-a-judge candidate selection; (3) ambiguity detection and multiple query generation for questions that are ambigious; (4) alignment-based filtering to remove misaligned NL-SQL pairs; and (5) final prompt-based filtering to remove any remaining low-quality instances (Figure~\ref{fig:pipeline}). We evaluate generation components by their contribution to SQL complexity and NL richness, and filtering components by their ability to retain high-quality examples while removing low-quality ones (validated against human annotations).

\subsection{SQL Query and NL Question Generation}
We adopt SQL-then-NL generation \cite{hu-etal-2023-importance, wu-etal-2021-data, lupidi2025source2synth}, which empirically increases both SQL and NL complexity (Table~\ref{tab:generation-ablations}). The LLM is prompted with a table schema to generate a SQL query, including foreign key relationships when present. Each query is executed against the database with up to three attempts to ensure validity before discarding.

The LLM is then prompted with the validated query and schema to generate four candidate NL questions via chain-of-thought prompting (Prompt~\ref{fig:prompt_nl_gen}). An LLM-as-a-judge selects the optimal candidate based on our rubric \cite{bai2022constitutional}. Requiring higher variation among candidates forces the model to produce more diverse linguistic options, improving selection quality (Section~\ref{gen-ablations}).

\subsubsection{Seeding Techniques}

To mitigate the inherent repetitiveness of synthetic data \citep{wang-etal-2023-self-instruct}, we interleave two seeding techniques to encourage the LLM to generate unique results (Table~\ref{tab:generation-ablations}). 

\paragraph{Persona Seeding.} Inspired by \citet{jandaghi-etal-2024-faithful, ge2025scalingsyntheticdatacreation}, for each table, we prompt the LLM to generate a realistic user persona with specific goals and example questions. We sample between two prompts to ensure a mix of common personas (e.g., "a software engineer") and creative ones (e.g., "an indie tabletop racing game designer"). This improves the complexity and diversity of generated questions while ensuring varied use cases (Table~\ref{tab:generation-ablations}, Prompt~\ref{fig:prompt_persona}, Prompt~\ref{fig:prompt_persona_creative}, Prompt~\ref{fig:prompt_persona_common}, Prompt~\ref{fig:prompt_nl_persona}).

\begin{figure}[ht]
    \centering
    \small
    \begin{tcolorbox}[
        colback=seedingblue!70!white, 
        colframe=seedingblue!50!black, 
        boxrule=0.5pt,
        boxsep=2pt,
        left=3pt,
        right=3pt,
        top=3pt,
        bottom=3pt
    ]
        \textbf{Persona:} TV producer revitalizing a reality show using 'The Amazing Race' data.
        
        \textbf{Goals:} Identify optimal timeslots and premiere dates.
        
        \textbf{Example:} How do fall vs. spring premiere ratings compare?
    \end{tcolorbox}
    \caption{Persona seeding example included in SQL generation prompts to introduce linguistic variety.}
    \label{fig:persona_example}
\end{figure}

\begin{figure*}[ht]
\begin{mdframed}
\noindent\textit{Persona (shortened):} Amanda Thompson, a travel blogger who uses 
storm record data to create safety guides for her readers.
\smallskip
\noindent\textit{NL:} ``What are the 5 regions that have experienced the most 
storms in the past 10 years, and what are their region names and codes, 
so I can warn my readers about these storm-prone areas?''
\begin{lstlisting}
SELECT `region`.`region name`, `region`.`region code`,
       COUNT(`storm`.`storm id`) AS `storm frequency`
FROM `region`
JOIN `affected region`
     ON `region`.`region id` = `affected region`.`region id`
JOIN `storm`
     ON `affected region`.`storm id` = `storm`.`storm id`
WHERE `storm`.`storm date` > DATE('now', '-10 years')
GROUP BY `region`.`region name`, `region`.`region code`
ORDER BY `storm frequency` DESC LIMIT 5;
\end{lstlisting}
\noindent The question is correctly grounded in the schema, joining \texttt{storm} and \texttt{region} through \texttt{affected\_region} to compute per-region storm counts, while the trailing context (``so I can warn my readers'') reflects how real users frame requests around goals rather than abstract retrieval.
\end{mdframed}
\caption{Example NL-SQL pair generated via persona seeding.}
\label{fig:example1}
\end{figure*}

\paragraph{Reserved Word Seeding.}
To align our synthetic SQL with human-authored distributions, we specify target reserved word counts sampled from BIRD's distribution \cite{li2023bird}, a widely-accepted challenging benchmark \cite{liu2025nl2sqlbugs, wretblad2024understanding, gao2024dailsql, pourreza2025chasesql}. The model achieves reasonable adherence (Pearson corelation $r=0.6$, median deviation of 3 words) and ablations confirm reserved word seeding improves SQL complexity despite small deviations (Table~\ref{tab:generation-ablations}, Prompt~\ref{fig:prompt_reserved}).

\subsubsection{Ambiguity Detection}
\label{subsubsec:ambig-pipeline}

After generating candidate NL-SQL pairs, we assign each example a continuous ambiguity score using a DistilBERT model fine-tuned on Ambrosia data \citep{ambrosia}. We define ambiguity as occurring whenever an NL question can be correctly interpreted as several distinct SQL queries.  Predicted scores $\hat{y}$ trigger two pipeline stages: (1) filtering removes extremely ambiguous questions, and (2) multi-interpretation generation (when $\hat{y} > 7$) produces alternative SQL queries for genuinely ambiguous examples using the prompt-based filters shown in Appendix Figures \ref{fig:prompt_filter_p1} and \ref{fig:prompt_filter_p2} to confirm genuine ambiguity and validate each alternative interpretation. Since this step may generate more than one SQL query for a single NL question, it also slightly increases the diversity of FALCON (see Appendix~\ref{app:ambiguity_model_training} for training details).  
Appendix Figures~\ref{fig:pipeline_walkthrough_ambig} and \ref{fig:pipeline_walkthrough_nonambig} provide a full pipeline walkthrough for an ambiguous and a non-ambiguous example respectively, and Appendix Figure~\ref{fig:example3} gives a further illustration.

\subsubsection{Alignment-Based Filtering}
\label{subsubsec:filtering}
Generative models frequently produce semantically plausible questions misaligned with SQL logic. We train an alignment prediction model on 9,000+ BIRD-derived examples (50\% systematically corrupted) using a novel hybrid metric combining semantic similarity (40\%) and schema entity matching (60\%) with corruption penalties. This scoring metric achieves Pearson correlation $r=0.70$ to human judgments (Appendix~\ref{app:alignment}). ModernBERT trained on this metric achieves MSE: 0.0203 and MAE: 0.1129, outperforming other models (Appendix Table~\ref{tab:alignment_models}) and filtering methods (Table~\ref{tab:quality_filtering}). This filtering step verifies that the generated query is faithfully grounded in the schema entities and relations the question implies, rejecting semantically plausible but ungrounded pairs.

\subsection{Multi-Interpretation Generation and Filtering}

Because ambiguity resolution is central to reliably constructing structured knowledge from natural language, we explicitly model and generate alternative interpretations rather than filtering ambiguous examples away. For queries flagged as ambiguous, we prompt the model to identify the ambiguity type \cite{ambrosia} and generate an alternative SQL interpretation (Prompt~\ref{fig:prompt_ambiguity}). The alternative is validated with an ambiguity-specific prompt filter (Prompt~\ref{fig:prompt_filter_p1}, ~\ref{fig:prompt_filter_p2}); the model has three attempts to produce an executable alternative that passes this filter, otherwise the example proceeds as unambiguous. 

All examples then undergo a final prompt-based filter verifying that: (1) the SQL validly interprets the NL, (2) the SQL is sensible and non-redundant, and (3) the NL resembles something a human would actually ask (Prompt~\ref{fig:prompt_final_filter}). While the reward model achieves high precision (Table~\ref{tab:quality_filtering}), it only reaches 71\% specificity and was trained solely on alignment, missing aligned examples with poor NL or SQL quality. Prompt-based filtering addresses these limitations and adapts to ambiguity-specific criteria beyond the reward model's scope. Examples passing all checks are added to the final dataset.

\section{Evaluation}

\subsection{Difficulty Metrics}
We measure NL Richness as a weighted combination of linguistic complexity features \citep{balyan2020applying, byrd2022predicting}: mean tokens before main verb (15\%), constituents per word (30\%), subordinate clauses (30\%), and coordinate clauses (20\%). SQL complexity is quantified using SQLGlot to parse query structures, with weighted features: words and columns (10\% each), and joins, sub-queries, operators, and splits (20\% each) \cite{yu-etal-2018-spider, li2023bird, vashistha2014measuring}.

\subsection{Human Evaluation}
We conduct human evaluation with 4 annotators using the rubric in Appendix~\ref{app:rubric}, which separately grades NL quality, SQL quality, and alignment to distinguish failure modes (e.g., inefficient queries vs. misaligned NL-SQL pairs). Each annotator independently scored 100 disjoint examples, with disagreements resolved through structured group discussion following standard adjudication protocol. 

\subsection{Downstream Task Evaluation}
We finetune Qwen 2.5 3B and 1.5B (Appendix ~\ref{app:qwen3btune}, ~\ref{app:qwen1btune}) on three training sets: (1) 5,500 synthetic examples, (2) 5,500 Spider examples, and (3) mixed data (95\% synthetic, 5\% Spider). Models are evaluated on our synthetic test set ($n=2{,}000$) and Spider test set ($n=2{,}147$) with consistent hyperparameters.

We employ three automatic metrics: \textbf{Exact Match (EM)}, direct string equivalence \cite{yu-etal-2018-spider}; \textbf{Fuzzy Score (FS)}, average string similarity; and \textbf{Soft Execution Accuracy (Soft-EX)}, F1 score of result set overlap \cite{bird_minidev}. Fuzzy metrics capture semantically equivalent queries with minor syntactic variations.  We conduct difficulty-stratified analysis across query complexity levels to characterize model performance on queries of varying difficulty.

\section{Ablations}

\subsection{Generation Order and Component Ablations}
\label{gen-ablations}
To maximize SQL complexity and NL richness, we ablated generation order (SQL-then-NL vs. NL-then-SQL), seeding methods (persona and reserved words), and LLM-as-a-judge selection strategies for NL: a Compare Judge (selecting the best candidate NL, Prompt~\ref{fig:prompt_judge}) and a Score Judge (iterative regeneration of NL until meeting a quality threshold, Prompt~\ref{fig:score_judge}). This NL candidate selection occurs before the final alignment-based filtering (Section~\ref{subsubsec:filtering}). 

Table~\ref{tab:generation-ablations} shows that SQL-then-NL outperforms the reverse order. Reserved word seeding improves SQL complexity while persona seeding improves NL richness. The Compare Judge yields the highest NL richness and unexpectedly improves SQL complexity by pruning simpler queries associated with low-quality questions. We therefore adopt SQL-then-NL generation, interleave both seeding methods, and use the Compare Judge in our final pipeline.

\begin{table}[h!]
  \centering
  \begin{tabular}{lcc}
    \hline
    \textbf{Configuration} & \textbf{SQL Comp.} & \textbf{NL Rich.} \\
    \hline
    \multicolumn{3}{l}{\textit{Generation Order}} \\
    SQL$\rightarrow$NL & \textbf{0.42} & \textbf{0.31} \\
    NL$\rightarrow$SQL & 0.40 & 0.21 \\
    \hline
    \multicolumn{3}{l}{\textit{Seeding \& Filtering (SQL$\rightarrow$NL)}} \\
    Baseline & 0.42 & 0.31 \\
    + Reserved Words & \textbf{0.48} & 0.31 \\
    + Personas & 0.42 & \textbf{0.34} \\
    + Compare Judge & 0.46 & \textbf{0.36} \\
    + Score Judge & 0.46 & 0.28 \\
    \hline    
  \end{tabular}
  \caption{Ablation results for generation order, seeding methods, and NL filtering strategies on SQL complexity and NL richness (n=1000) using Qwen-2.5-3B.}
  \label{tab:generation-ablations}
\end{table}

\subsection{Filtering Ablation}
\label{filtering-ablations}

\paragraph{Filtering Method Comparison}

We evaluate our reward model against three filtering methods on human-annotated data (n=181, 84\% aligned): \textbf{(1) Teacher-based}, where a larger model (Llama-70B) regenerates SQL (k=3 tries), discarding non-reproducible examples; \textbf{(2) Source2Synth} \citet{lupidi2025source2synth}, which fine-tunes the generator (Qwen-2.5-3B) on unfiltered data and rejects examples it cannot reproduce in k=3 attempts. This performed poorly, suggesting limited generalization to smaller models; and \textbf{(3) Prompt-based LLM-as-judge}, which uses Llama-3-70B to score alignment via chain-of-thought reasoning.

Our reward model substantially outperforms baselines (Table~\ref{tab:quality_filtering}), achieving 94\% F1 and 71\% specificity at threshold 0.6. It reduces misaligned examples from 16\% to 4\% while retaining 92.81\% of correct examples (8-9× higher recall than competing methods). Filtered examples' SQL and NL complexity metrics match BIRD benchmarks, confirming our filtering preserves difficulty without distributional bias (Appendix~\ref{tab:llm_experiments}).

\begin{table}[htbp]
\centering

\resizebox{\columnwidth}{!}{
\begin{tabular}{lccc}
\toprule
\textbf{Method} & \textbf{F1} & \textbf{Specificity} & \textbf{Recall} \\
\midrule
Reward Model          & \textbf{93.73} & 71.43          & 92.81 \\
Prompt Filter         & 91.54          & 28.57          & \textbf{95.42} \\
LLaMA Teacher         & 17.65          & 92.86          &  9.80 \\
Source2Synth          &  7.55          & \textbf{100.00}&  3.92 \\
\bottomrule
\end{tabular}
}
\caption{Filtering results across different methods. F1 measures overall filtering performance. Specificity shows the percentage of low-quality examples correctly filtered out. Recall shows the percentage of high-quality examples retained. Results are shown at the best threshold for each model.}
\label{tab:quality_filtering}
\end{table}

\subsection{Ambiguity Ablation}

We evaluate the ambiguity pipeline on 100 human-annotated examples from our final dataset. For each example, human annotators judged whether the query was genuinely ambiguous and whether the generated alternative SQL queries were valid. An example was counted as a true positive only when both conditions held. Those judgments were aggregated into a confusion matrix from which we compute accuracy, precision, recall, F1, and specificity. Table~\ref{tab:ambiguity_detection_performance} shows 94\% accuracy with 98.9\% specificity with  28.6\% recall. This conservative behavior is intentional: false positives generate noisy multi-interpretation pairs, while false negatives simply omit optional enhancements. High specificity ensures triggered multi-interpretation generation produces genuine alternatives. 

\begin{table}[h!]
\centering
\small
\begin{tabular}{lccccc}
\toprule
\textbf{Accuracy} & \textbf{Precision} & \textbf{Recall} & \textbf{F1} & \textbf{Specificity} \\
\midrule
94.0\% & 66.7\% & 28.6\% & 40.0\% & 98.9\% \\
\bottomrule
\end{tabular}
\caption{End-to-end ambiguity pipeline performance on human-annotated samples from final synthetic dataset (n=100), evaluating both detection accuracy and validity of generated alternative queries.}
\label{tab:ambiguity_detection_performance}
\end{table}

\section{Results}

Human evaluation with four annotators (100 examples each using the rubric in Appendix~\ref{app:rubric}) shows consistently high quality across all models (Table~\ref{tab:llm_quality}), with SQL, NL, and alignment scores all exceeding 0.92. Quality holds across seed base databases (Appendix~\ref{tab:human_eval_cross_databases}), and complexity remains consistent across models without declining after filtering—in fact, it improves (Appendix~\ref{tab:llm_experiments}).

\begin{table}[t]
  \centering
  \begin{tabular}{lcccc}
    \hline
    \textbf{Model} & \textbf{SQL} & \textbf{NL} & \textbf{Alignment} \\
    \hline
    Qwen-3B  & 0.96 &  0.95 & 0.94  \\
    Llama-8B  & 0.92 & 0.94 & 0.95 \\
    Qwen-32B & 0.98 & 0.97 & 0.93  \\
    Llama-70B & 0.93 & 0.97 & 0.94 \\
    \hline    
  \end{tabular}
  \caption{Human evaluation scores across model sizes for the FALCON-generated data that was seeded with WikiSQL and filtered. All models were evaluated according to our rubric. We used the following models: Qwen-2.5-3B, Llama-3.1-8B, Qwen-3-32B, and Llama-3.3-70B.}
  \label{tab:llm_quality}
\end{table}

Table~\ref{tab:dataset_comp} shows that our FALCON-generated datasets outperform the original datasets in terms of SQL complexity and NL richness, despite using the same table information. With the exception of the WikiSQL-seeded pipeline's SQL complexity, the FALCON-generated data is comparable, if not exceeding the complexity of human generated data in BIRD.

\begin{table}[t]
  \centering
  \resizebox{\columnwidth}{!}{
  \begin{tabular}{lccc}
    \hline
    \textbf{Dataset} & \textbf{SQL Comp.} & \textbf{NL Rich.}\\
    \hline
    Spider Seeded FALCON & \textbf{0.43} & \textbf{0.39}  \\
    WikiSQL Seeded FALCON & 0.28 & \textbf{0.39}\\
    BIRD & 0.42 & 0.31  \\
    Spider  & 0.38 & 0.29 \\
    WikiSQL & 0.20 & 0.28 \\
    \hline    
  \end{tabular}
  }
  \caption{Comparison of the SQL complexity and NL richness of multiple datasets with the most complex scores in bold. Spider seeded is FALCON seeded on spider data. WikiSQL seeded is FALCON seeded on WikiSQL.}
  \label{tab:dataset_comp}
\end{table}

\begin{figure}[h!]
    \centering
    \includegraphics[width=1\linewidth]{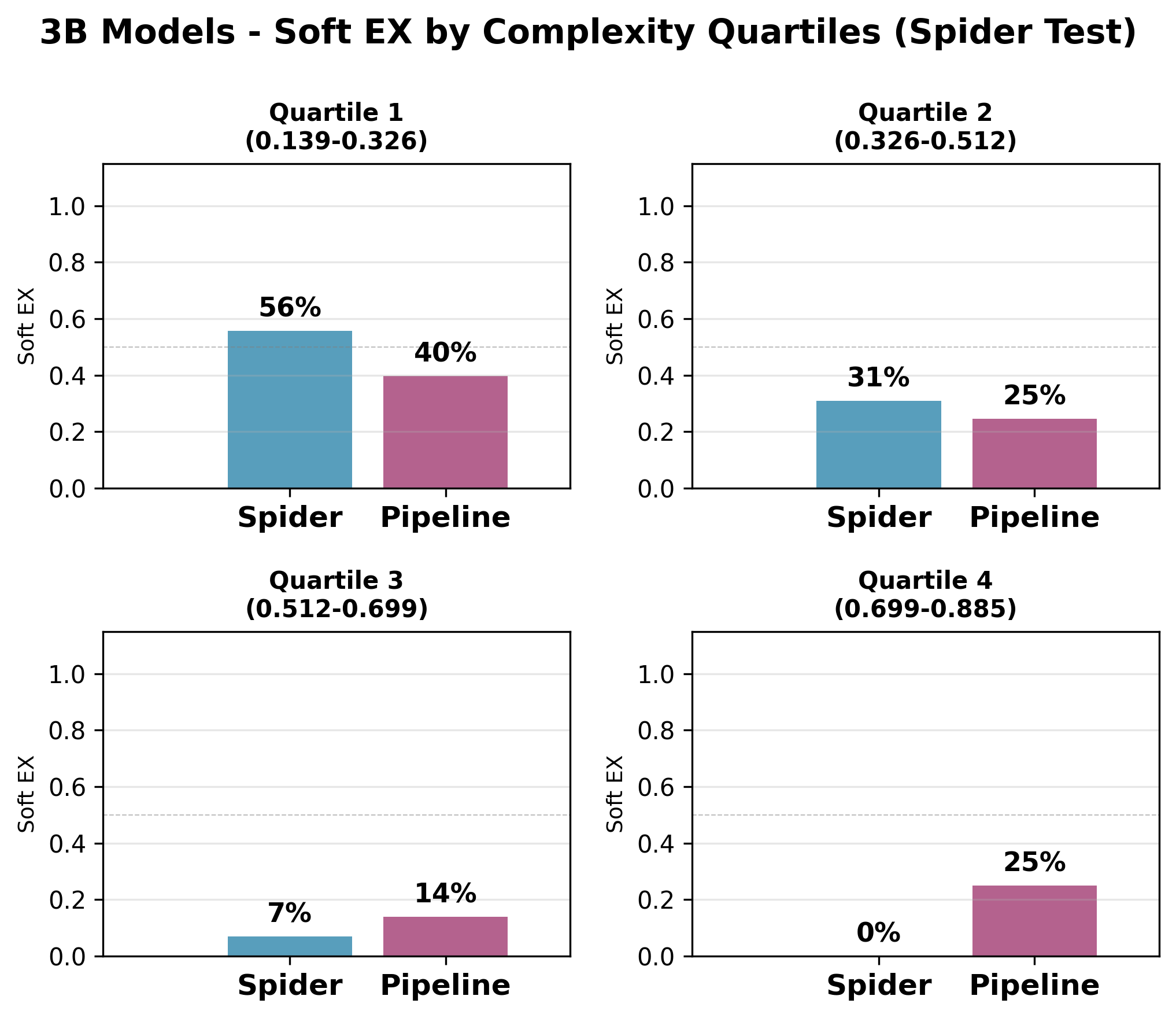}
    \caption{A comparison of soft execution accuracy for a model finetuned on Spider and the FALCON-generated data for different complexities.}
    \label{fig:quartiles}
\end{figure}

\begin{figure}[h!]
    \centering
    \includegraphics[width=1\linewidth]{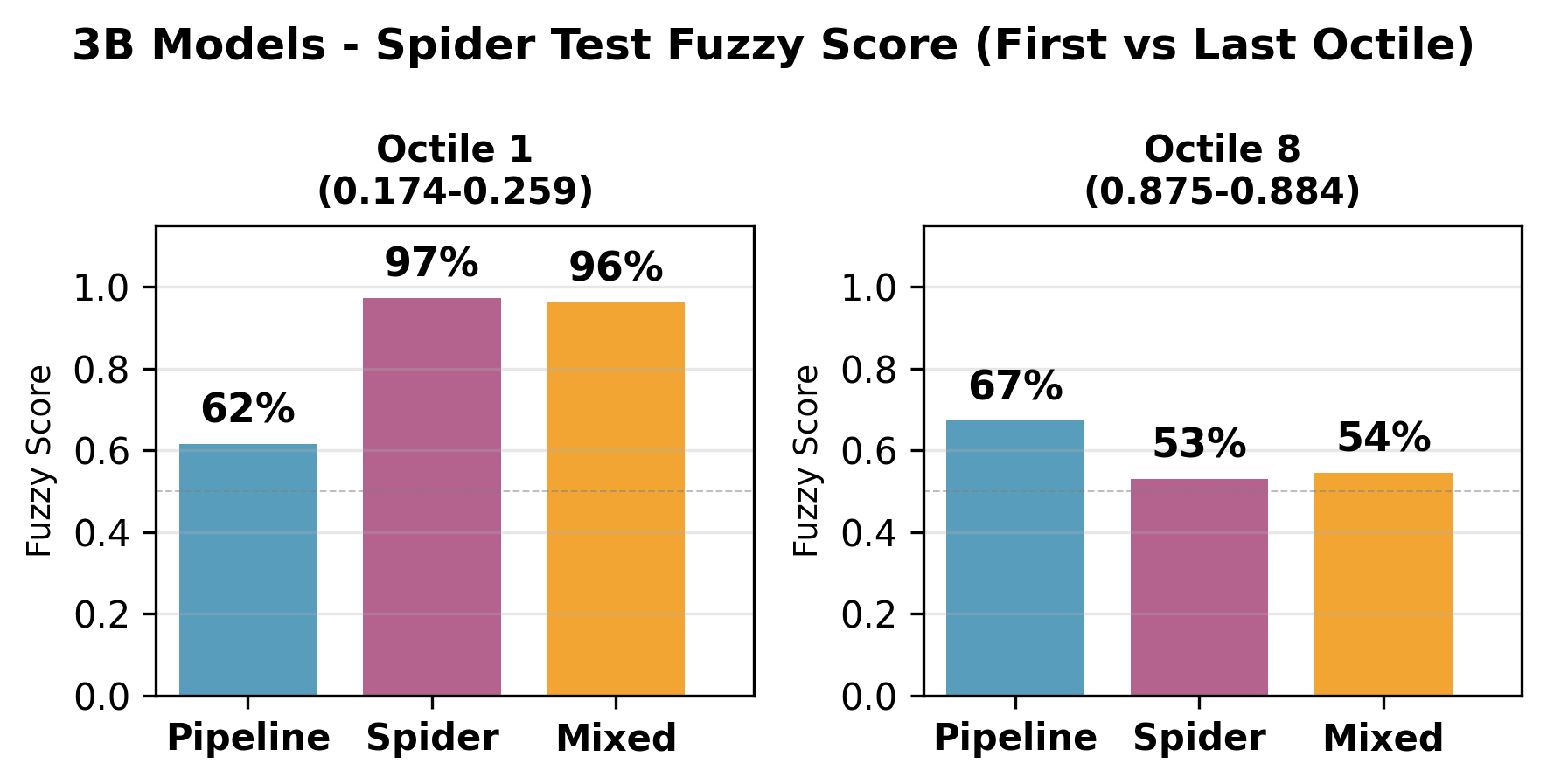}
    \caption{The fuzzy scores for a model finetuned on FALCON, Spider, and a mix of both.}
    \label{fig:fuzzy}
\end{figure}

Our downstream results validate FALCON's core objective: generating challenging training data. Difficulty-stratified analysis demonstrates that FALCON-trained models significantly outperform Spider-trained models as query complexity increases, with the performance advantage growing in harder quartiles (Figure~\ref{fig:quartiles}).
Spider-trained models achieve 36\% EM on Spider but only 0.05\% on FALCON, while FALCON-trained models achieve 13.8\% on FALCON but 0.8\% on Spider (Table~\ref{tab:cross_dataset_eval_3b}), reflecting Spider's over-representation of simpler queries (Appendix~\ref{app:complexity_distribution}). The low raw EM on Spider is consistent with known string match underestimation when training and test distributions diverge significantly. Our error analysis finds approximately a third of incorrect predictions are acceptable variations undetected by string match, likely amplified here relative to Spider-trained baselines due to the larger distribution gap (Appendix~\ref{app:error_analysis}). When supplemented with 5\% Spider data, mixed training achieves 27\% EM on Spider and 13.6\% on FALCON (raw string match). Figure~\ref{fig:fuzzy} shows that real data seeding primarily benefits easier queries (first octile) while FALCON-only training maintains advantages on harder queries (last octile).
The most common genuine failure modes are spurious predicate 
hallucination and result boundary errors, attributable to the 
overrepresentation of complex queries in our training data; 
mixing with 5\% real data is expected to reduce these artifacts 
(Section~\ref{app:error_analysis_main}).

\subsection{Error Analysis}
\label{app:error_analysis_main}

We conducted a failure mode analysis on 100 randomly sampled incorrect predictions from the Qwen 2.5-3B model finetuned on FALCON's generations, evaluated on the Spider test set, identifying nine error categories ranging from acceptable variations to genuine failures such as spurious predicate hallucination (adding a WHERE cause where none is needed) and result boundary errors (returning too much or too little) (full taxonomy and examples in~\ref{app:error_analysis}). Table~\ref{tab:error_analysis_results} shows 
the distribution of error types in the annotated sample.

\begin{table}[h]
\centering
\small
\begin{tabular}{lcc}
\toprule
\textbf{Error Type} & \textbf{Count} & \textbf{\%} \\
\midrule
Acceptable variation          & 33 & 33.0 \\
Spurious predicate hallucination & 20 & 20.0 \\
Result boundary error         & 18 & 18.0 \\
Schema entity confusion       & 10 & 10.0 \\
Logical negation failure      &  4 &  4.0 \\
Degenerate join structure     &  4 &  4.0 \\
Cross-row constraint flattening &  4 &  4.0 \\
Hardcoded quantifier substitution &  4 &  4.0 \\
NL misinterpretation          &  3 &  3.0 \\
\midrule
\textbf{Total}                & 100 & 100.0 \\
\bottomrule
\end{tabular}
\caption{Distribution of error types across 100 sampled incorrect 
predictions from the model finetuned on FALCON-generated data, 
evaluated on the Spider test set.}
\label{tab:error_analysis_results}
\end{table}

\paragraph{Analysis.} From this analysis we draw three conclusions. First, approximately one third of flagged errors are acceptable variations that would not mislead a real user, meaning our reported Spider accuracy is meaningfully underestimated. This is a known limitation of both string match and execution match metrics. This underestimation likely affects our model more than Spider-trained baselines, since style-based acceptable variations are more probable when training and test distributions diverge substantially. Second, spurious predicate hallucination and result boundary errors are the most common genuine failure modes, which we attribute to the overrepresentation of complex 
SQL queries in our dataset. We anticipate that mixing the 
training distribution with simpler data has the most impact 
here by reducing these artifacts, which is why we recommend 
seeding with 5\% existing data (ex. from Spider or WikiSQL) to ensure simpler queries are represented. Third, 
schema entity confusion is more likely a consequence of how 
the model processes condensed schema information under context 
pressure than of our dataset composition: when training the 
model, we compressed schema information for all tables into 
the prompt, requiring the model to resolve column and table 
references under significant context pressure, independently 
of the underlying NL-to-SQL pairs.

\begin{table}[h]
\centering
\setlength{\tabcolsep}{4pt}
\begin{tabular}{lccc}
\toprule
\textbf{Train $\rightarrow$ Test} & \textbf{EM (\%)} & \textbf{FS} & \textbf{Soft-EX} \\
\midrule
Spider on Spider  & \textbf{35.95} & \textbf{0.89} & \textbf{0.42} \\
Spider on Pipe    &  0.05 & 0.69 & 0.18 \\
FALCON on Spider    &  0.80$^{\dagger}$ & 0.68 & 0.32 \\
FALCON on Pipe      & \textbf{13.80} & \textbf{0.83} & \textbf{0.48} \\
Mixed on Spider   & 27.10 & 0.84 & -- \\
Mixed on Pipe     & 13.60 & \textbf{0.83} & -- \\
\bottomrule
\end{tabular}
\caption{Cross-dataset evaluation on Qwen 2.5-3B. In-distribution rows serve as baselines contextualizing the cross-dataset results. Mixed is the intended deployment configuration. EM = Exact Match, FS = Fuzzy Score, Soft-EX = Soft Execution Accuracy. 
$^{\dagger}$Raw EM underestimates performance due to acceptable variations undetected by string match (Appendix~\ref{app:error_analysis}).}
\label{tab:cross_dataset_eval_3b}
\end{table}

\subsection{Cost}
FALCON requires no external API access, enabling organizations to generate training data locally without per-query costs or privacy exposure. Using on average 5 LLM calls per example across 5 pipeline stages, generating 5,500 examples requires approximately 34.47 million input tokens and 5.13 million output tokens. At GPT-5.5 pricing, an equivalent run would cost approximately \$326, versus \$20 for an open model on a commodity GPU (Vast.ai, \$0.15/hr), an
over 16$\times$ reduction, before accounting for the privacy implications of 
transmitting proprietary schemas to third-party APIs (Appendix~\ref{app:cost}).

\section{Conclusion}

We present FALCON, a model- and database-agnostic framework for generating high-complexity NL-to-SQL data at low cost using compact open models. Through reserved-word seeding, persona-based prompting, and alignment-based filtering, FALCON generates data exceeding existing benchmarks in SQL complexity and natural language richness while maintaining high quality (>0.92 human evaluation across all models). Our ambiguity handling addresses a critical gap in synthetic generation, producing diverse data reflecting real-world linguistic challenges.

Models trained on FALCON-generated data outperform Spider-trained models as query complexity increases. FALCON's ambiguity-aware, schema-grounded approach to NL-to-SQL generation offers a template for producing training data for natural language interfaces to structured knowledge more broadly, and its local, model-agnostic design keeps this accessible to organizations without reliance on frontier APIs.

\section{Limitations}

FALCON is intentionally designed to target the complex end of the query distribution, where existing datasets and template-based methods already provide adequate coverage of simpler queries. This focus leads to an underrepresentation of simpler queries, as shown in Figure~\ref{fig:complexity_dist}, and models trained solely on FALCON data achieve only 0.8\% EM on Spider, requiring 5--10\% real data supplementation for broad applicability. Ambiguity detection achieves high specificity (98.9\%) but low recall (28.6\%), missing many genuinely ambiguous queries.

FALCON currently supports SQLite only; extension to other dialects 
requires dialect-specific reserved word distributions and syntax 
constraints. Evaluation is limited to public benchmarks (Spider, 
WikiSQL, BIRD), and production databases may exhibit complexity 
dimensions not captured academically. Resource constraints limited 
downstream evaluation to compact models, 
though these suffice for high-quality generation ($>$0.92 human 
evaluation scores). Generated data inherits domain coverage from seed 
databases.

\section*{Ethical Considerations}

Domain coverage reflects source databases (Spider and WikiSQL), potentially inheriting domain-specific biases. The database-agnostic design enables organizations to generate data for their specific use cases. AI tools were used to aid in non-essential writing and code generation.

\section*{Acknowledgments}
The authors would like to thank the anonymous reviewers for their feedback. We would also like to thank Shriya Sravani Yellapragada for her early work on this research. Additional thanks to Beth Ann Hockey and Jeffrey Flanigan for their advice and support.

\bibliography{custom}

\appendix

\section{Appendix}
\label{sec:appendix}

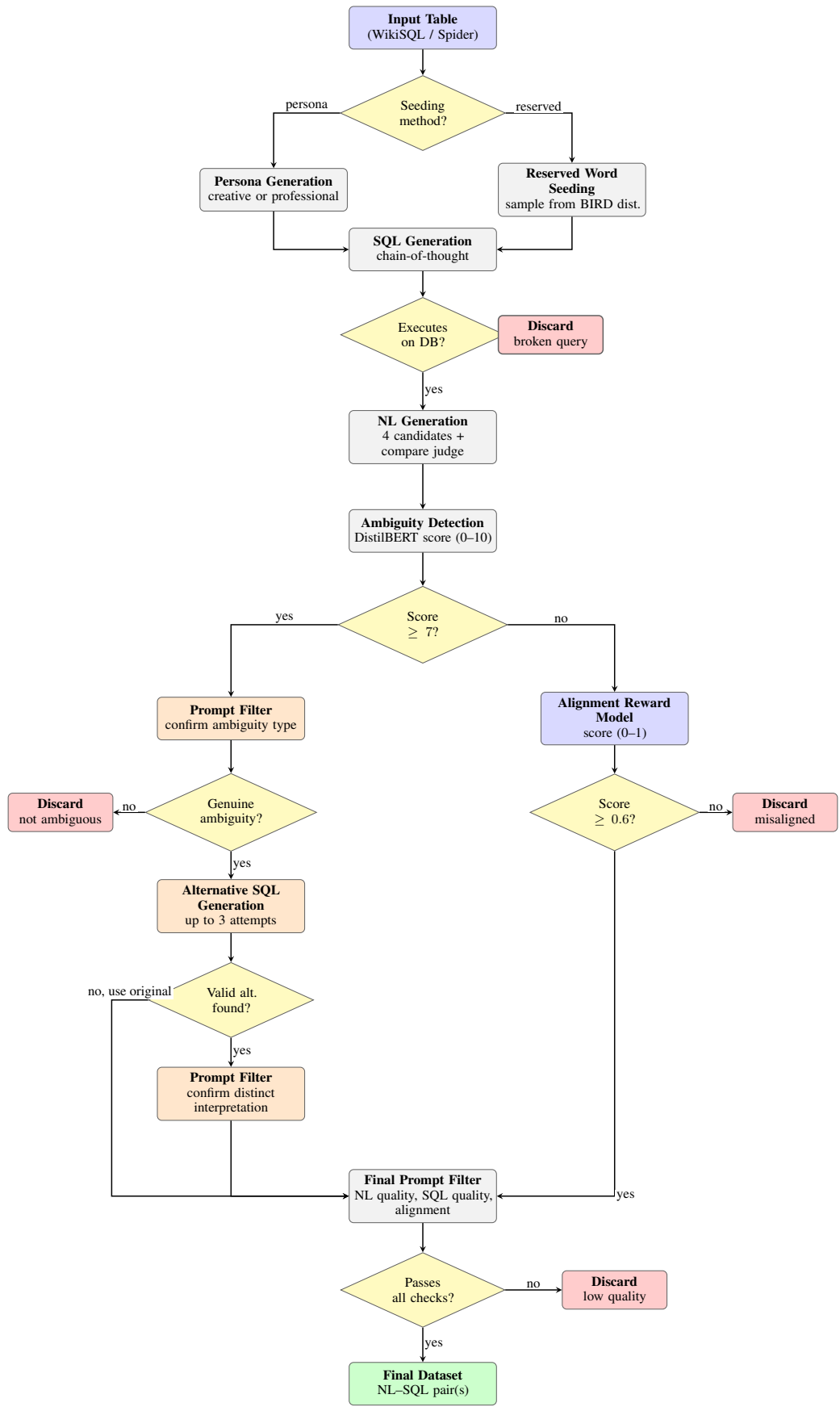
\begin{figure*}[h]
\centering
\resizebox{0.85\textwidth}{!}{%
\begin{tikzpicture}[
    node distance=1.2cm and 3.5cm,
    box/.style={
        rectangle,
        rounded corners=3pt,
        draw=gray!75!black,
        fill=gray!10!white,
        text width=3.2cm,
        minimum height=1.0cm,
        align=center,
        font=\small
    },
    decision/.style={
        diamond,
        draw=gray!75!black,
        fill=yellow!30!white,
        text width=1.8cm,
        minimum height=1.0cm,
        align=center,
        font=\small,
        aspect=2.2
    },
    output/.style={
        rectangle,
        rounded corners=3pt,
        draw=gray!75!black,
        fill=green!20!white,
        text width=3.2cm,
        minimum height=1.0cm,
        align=center,
        font=\small
    },
    discard/.style={
        rectangle,
        rounded corners=3pt,
        draw=gray!75!black,
        fill=red!20!white,
        text width=2.2cm,
        minimum height=0.9cm,
        align=center,
        font=\small
    },
    ambig/.style={
        rectangle,
        rounded corners=3pt,
        draw=gray!75!black,
        fill=orange!20!white,
        text width=3.2cm,
        minimum height=1.0cm,
        align=center,
        font=\small
    },
    reward/.style={
        rectangle,
        rounded corners=3pt,
        draw=gray!75!black,
        fill=blue!15!white,
        text width=3.2cm,
        minimum height=1.0cm,
        align=center,
        font=\small
    },
    arrow/.style={->, >=stealth, thick},
    label/.style={font=\footnotesize, fill=white, inner sep=1pt}
]

\node[box, fill=blue!15!white] (table)    at (0,  0)    
    {\textbf{Input Table}\\(WikiSQL / Spider)};
\node[decision]               (seed)     at (0, -2.0)  
    {Seeding\\method?};
\node[box]                    (sqlgen)   at (0, -5.2)  
    {\textbf{SQL Generation}\\chain-of-thought};
\node[decision]               (execok)   at (0, -7.2)  
    {Executes\\on DB?};
\node[box]                    (nlgen)    at (0, -9.6)  
    {\textbf{NL Generation}\\4 candidates +\\compare judge};
\node[box]                    (ambigdet) at (0, -11.8) 
    {\textbf{Ambiguity Detection}\\DistilBERT score (0--10)};
\node[decision]               (ambigth)  at (0, -14.0) 
    {Score\\$\geq 7$?};

\node[ambig]     (pf1)        at (-4.5, -16.2) 
    {\textbf{Prompt Filter}\\confirm ambiguity type};
\node[decision]  (genuineq)   at (-4.5, -18.4) 
    {Genuine\\ambiguity?};
\node[discard]   (disc2)      at (-8.5, -18.4) 
    {\textbf{Discard}\\not ambiguous};
\node[ambig]     (altgen)     at (-4.5, -20.6) 
    {\textbf{Alternative SQL\\Generation}\\up to 3 attempts};
\node[decision]  (altok)      at (-4.5, -22.8) 
    {Valid alt.\\found?};
\node[ambig]     (pf2)        at (-4.5, -25.0) 
    {\textbf{Prompt Filter}\\confirm distinct\\interpretation};

\node[reward]    (reward)     at (4.5, -16.2)  
    {\textbf{Alignment Reward\\Model}\\score (0--1)};
\node[decision]  (rewardth)   at (4.5, -18.4)  
    {Score\\$\geq 0.6$?};
\node[discard]   (disc3)      at (8.5, -18.4)  
    {\textbf{Discard}\\misaligned};

\node[box]       (finalfilt)  at (0, -27.4)    
    {\textbf{Final Prompt Filter}\\NL quality, SQL quality,\\alignment};
\node[decision]  (finalok)    at (0, -29.6)    
    {Passes\\all checks?};
\node[discard]   (disc4)      at (4.5, -29.6)  
    {\textbf{Discard}\\low quality};
\node[output]    (dataset)    at (0, -31.8)    
    {\textbf{Final Dataset}\\NL--SQL pair(s)};

\node[box] (persona)  at (-3.5, -3.8) 
    {\textbf{Persona Generation}\\creative or professional};
\node[box] (reserved) at ( 3.5, -3.8) 
    {\textbf{Reserved Word\\Seeding}\\sample from BIRD dist.};

\draw[arrow] (table)    -- (seed);
\draw[arrow] (seed)     -| node[label, near start, above] {persona}  
    (persona);
\draw[arrow] (seed)     -| node[label, near start, above] {reserved} 
    (reserved);
\draw[arrow] (persona)  |- (sqlgen);
\draw[arrow] (reserved) |- (sqlgen);
\draw[arrow] (sqlgen)   -- (execok);
\draw[arrow] (execok)   -- node[label, right] {yes} (nlgen);
\draw[arrow] (execok)   -- node[label, above] {no}  
    ++(3.0, 0) node[discard] (disc1) 
    {\textbf{Discard}\\broken query};
\draw[arrow] (nlgen)    -- (ambigdet);
\draw[arrow] (ambigdet) -- (ambigth);

\draw[arrow] (ambigth)  -| node[label, near start, above] {yes} (pf1);
\draw[arrow] (pf1)      -- (genuineq);
\draw[arrow] (genuineq) -- node[label, above] {no}  (disc2);
\draw[arrow] (genuineq) -- node[label, right] {yes} (altgen);
\draw[arrow] (altgen)   -- (altok);
\draw[arrow] (altok)    -- node[label, right] {yes} (pf2);
\draw[arrow] (altok)    -- node[label, above] {no, use original} 
    ++(-2.8, 0) coordinate (altno) 
    |- (finalfilt);
\draw[arrow] (pf2)      |- (finalfilt);

\draw[arrow] (ambigth)  -| node[label, near start, above] {no} (reward);
\draw[arrow] (reward)   -- (rewardth);
\draw[arrow] (rewardth) -- node[label, above] {no}  (disc3);
\draw[arrow] (rewardth) |- node[label, right] {yes} (finalfilt);

\draw[arrow] (finalfilt) -- (finalok);
\draw[arrow] (finalok)   -- node[label, above] {no}  (disc4);
\draw[arrow] (finalok)   -- node[label, right] {yes} (dataset);

\end{tikzpicture}%
}
\caption{FALCON pipeline flowchart. Yellow diamonds indicate 
decision points. Orange boxes indicate ambiguity handling steps. 
Blue/purple boxes indicate alignment filtering. Green indicates 
the final dataset output. Red indicates discarded examples.}
\label{fig:pipeline_flowchart}
\end{figure*}

\subsection{Detailed Pipeline Overview}
\label{app:complex}
\begin{figure*}[h!]
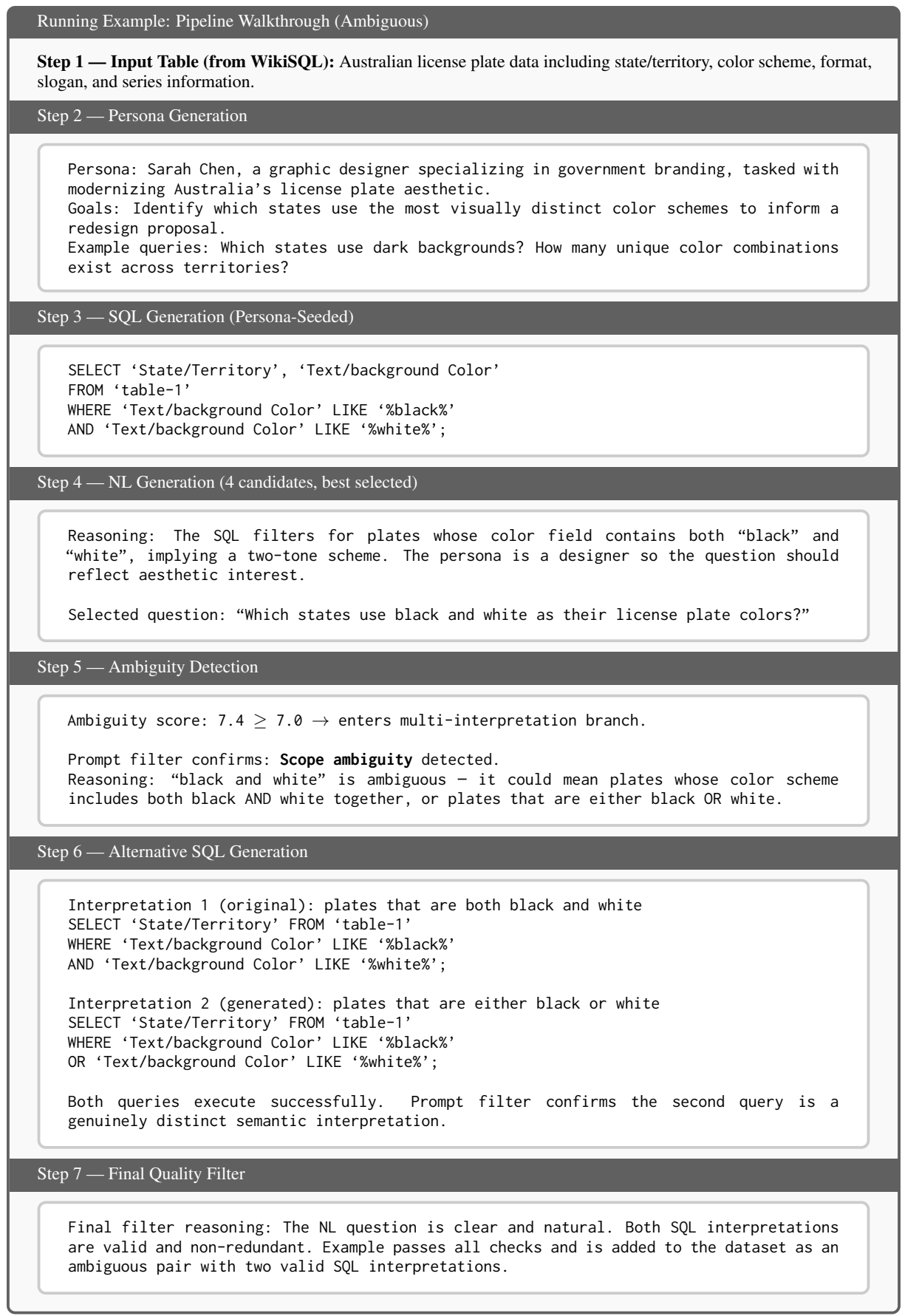

\centering
\small
\begin{tcolorbox}[
    colback=gray!5!white,
    colframe=gray!75!black,
    title=Running Example: Pipeline Walkthrough (Ambiguous),
    fontupper=\small,
    width=\textwidth
]
\textbf{Step 1 --- Input Table (from WikiSQL):} Australian license 
plate data including state/territory, color scheme, format, slogan, 
and series information.
\tcbsubtitle{Step 2 --- Persona Generation}
\begin{tcolorbox}[colback=white, colframe=gray!40!white, 
fontupper=\small\ttfamily]
Persona: Sarah Chen, a graphic designer specializing in government 
branding, tasked with modernizing Australia's license plate aesthetic.\\
Goals: Identify which states use the most visually distinct color 
schemes to inform a redesign proposal.\\
Example queries: Which states use dark backgrounds? How many unique 
color combinations exist across territories?
\end{tcolorbox}
\tcbsubtitle{Step 3 --- SQL Generation (Persona-Seeded)}
\begin{tcolorbox}[colback=white, colframe=gray!40!white, 
fontupper=\small\ttfamily]
SELECT `State/Territory', `Text/background Color'\\
FROM `table-1'\\
WHERE `Text/background Color' LIKE `\%black\%'\\
~~~AND `Text/background Color' LIKE `\%white\%';
\end{tcolorbox}
\tcbsubtitle{Step 4 --- NL Generation (4 candidates, best selected)}
\begin{tcolorbox}[colback=white, colframe=gray!40!white, 
fontupper=\small\ttfamily]
Reasoning: The SQL filters for plates whose color field contains 
both ``black'' and ``white'', implying a two-tone scheme. The persona 
is a designer so the question should reflect aesthetic interest.\\
\\
Selected question: ``Which states use black and white as their 
license plate colors?''
\end{tcolorbox}
\tcbsubtitle{Step 5 --- Ambiguity Detection}
\begin{tcolorbox}[colback=white, colframe=gray!40!white, 
fontupper=\small\ttfamily]
Ambiguity score: 7.4 $\geq$ 7.0 $\rightarrow$ enters 
multi-interpretation branch.\\
\\
Prompt filter confirms: \textbf{Scope ambiguity} detected.\\
Reasoning: ``black and white'' is ambiguous --- it could mean plates 
whose color scheme includes both black AND white together, or plates 
that are either black OR white.
\end{tcolorbox}
\tcbsubtitle{Step 6 --- Alternative SQL Generation}
\begin{tcolorbox}[colback=white, colframe=gray!40!white, 
fontupper=\small\ttfamily]
Interpretation 1 (original): plates that are both black and white\\
SELECT `State/Territory' FROM `table-1'\\
WHERE `Text/background Color' LIKE `\%black\%'\\
~~~AND `Text/background Color' LIKE `\%white\%';\\
\\
Interpretation 2 (generated): plates that are either black or white\\
SELECT `State/Territory' FROM `table-1'\\
WHERE `Text/background Color' LIKE `\%black\%'\\
~~~OR `Text/background Color' LIKE `\%white\%';\\
\\
Both queries execute successfully. Prompt filter confirms the 
second query is a genuinely distinct semantic interpretation.
\end{tcolorbox}
\tcbsubtitle{Step 7 --- Final Quality Filter}
\begin{tcolorbox}[colback=white, colframe=gray!40!white, 
fontupper=\small\ttfamily]
Final filter reasoning: The NL question is clear and natural. 
Both SQL interpretations are valid and non-redundant. 
Example passes all checks and is added to the dataset as an 
ambiguous pair with two valid SQL interpretations.
\end{tcolorbox}
\end{tcolorbox}
\caption{Step-by-step pipeline walkthrough using a persona-seeded 
ambiguous example. Intermediate outputs are simplified for 
illustration purposes.}
\label{fig:pipeline_walkthrough_ambig}
\end{figure*}

\begin{figure*}[h!]
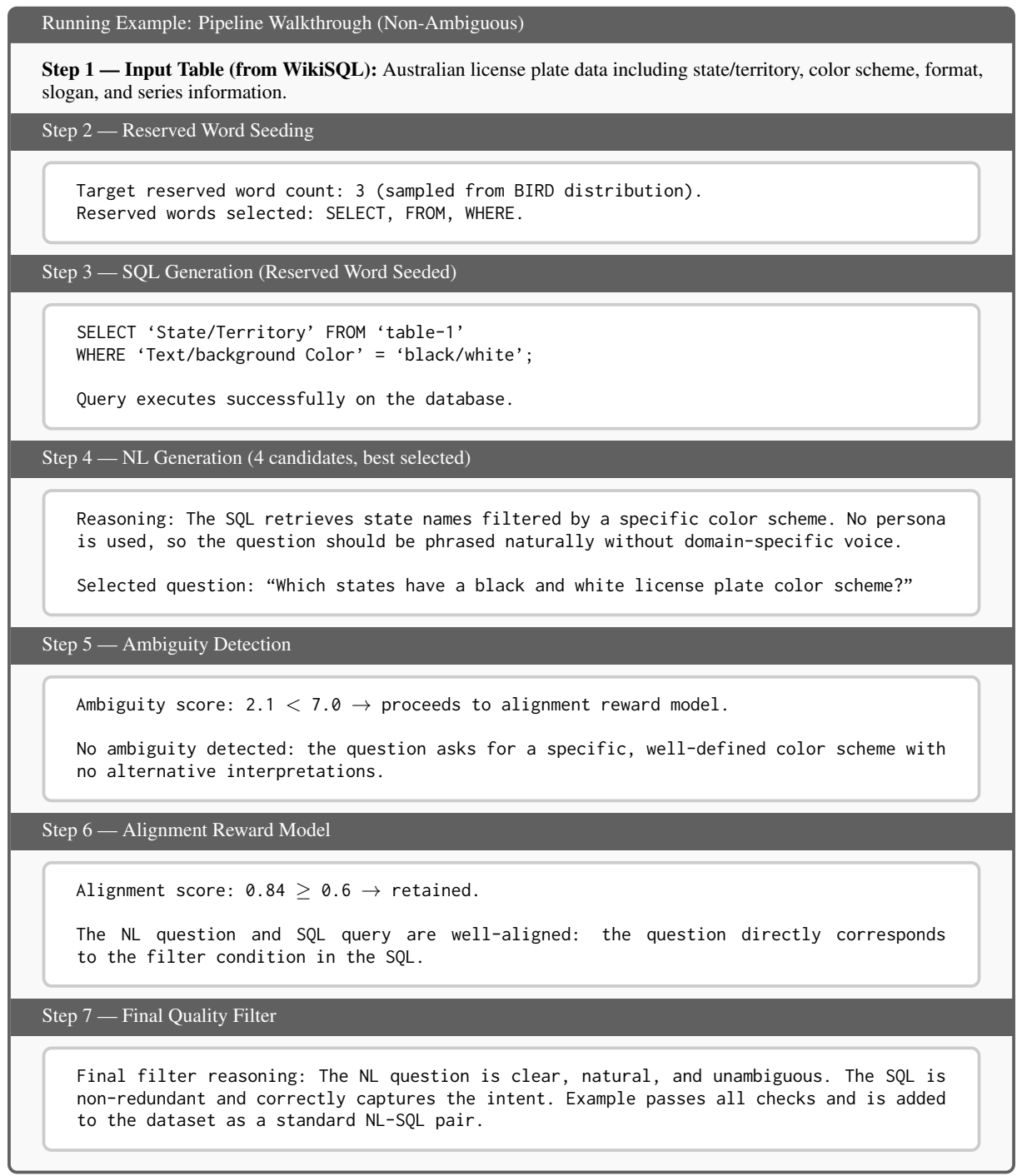

\centering
\small
\begin{tcolorbox}[
    colback=gray!5!white,
    colframe=gray!75!black,
    title=Running Example: Pipeline Walkthrough (Non-Ambiguous),
    fontupper=\small,
    width=\textwidth
]
\textbf{Step 1 --- Input Table (from WikiSQL):} Australian license 
plate data including state/territory, color scheme, format, slogan, 
and series information.
\tcbsubtitle{Step 2 --- Reserved Word Seeding}
\begin{tcolorbox}[colback=white, colframe=gray!40!white, 
fontupper=\small\ttfamily]
Target reserved word count: 3 (sampled from BIRD distribution).\\
Reserved words selected: SELECT, FROM, WHERE.
\end{tcolorbox}
\tcbsubtitle{Step 3 --- SQL Generation (Reserved Word Seeded)}
\begin{tcolorbox}[colback=white, colframe=gray!40!white, 
fontupper=\small\ttfamily]
SELECT `State/Territory' FROM `table-1'\\
WHERE `Text/background Color' = `black/white';\\
\\
Query executes successfully on the database.
\end{tcolorbox}
\tcbsubtitle{Step 4 --- NL Generation (4 candidates, best selected)}
\begin{tcolorbox}[colback=white, colframe=gray!40!white, 
fontupper=\small\ttfamily]
Reasoning: The SQL retrieves state names filtered by a specific 
color scheme. No persona is used, so the question should be 
phrased naturally without domain-specific voice.\\
\\
Selected question: ``Which states have a black and white license 
plate color scheme?''
\end{tcolorbox}
\tcbsubtitle{Step 5 --- Ambiguity Detection}
\begin{tcolorbox}[colback=white, colframe=gray!40!white, 
fontupper=\small\ttfamily]
Ambiguity score: 2.1 $<$ 7.0 $\rightarrow$ proceeds to alignment 
reward model.\\
\\
No ambiguity detected: the question asks for a specific, 
well-defined color scheme with no alternative interpretations.
\end{tcolorbox}
\tcbsubtitle{Step 6 --- Alignment Reward Model}
\begin{tcolorbox}[colback=white, colframe=gray!40!white, 
fontupper=\small\ttfamily]
Alignment score: 0.84 $\geq$ 0.6 $\rightarrow$ retained.\\
\\
The NL question and SQL query are well-aligned: the question 
directly corresponds to the filter condition in the SQL.
\end{tcolorbox}
\tcbsubtitle{Step 7 --- Final Quality Filter}
\begin{tcolorbox}[colback=white, colframe=gray!40!white, 
fontupper=\small\ttfamily]
Final filter reasoning: The NL question is clear, natural, and 
unambiguous. The SQL is non-redundant and correctly captures 
the intent. Example passes all checks and is added to the 
dataset as a standard NL-SQL pair.
\end{tcolorbox}
\end{tcolorbox}
\caption{Step-by-step pipeline walkthrough using a reserved-word-seeded 
non-ambiguous example. Intermediate outputs are simplified for 
illustration purposes.}
\label{fig:pipeline_walkthrough_nonambig}
\end{figure*}

We demonstrate the pipeline step by step using a running example. 
Note that all examples are simplified and made up for illustration 
purposes only; they are not taken directly from the pipeline output. A 
detailed diagram of the full pipeline is available in 
Figure~\ref{fig:pipeline_flowchart}, and step-by-step walkthroughs 
showing intermediate outputs for both an ambiguous and a 
non-ambiguous example are shown in 
Figures~\ref{fig:pipeline_walkthrough_ambig} 
and~\ref{fig:pipeline_walkthrough_nonambig} respectively.

\textbf{Table:}
The pipeline begins with an existing dataset of table schemas, 
such as WikiSQL, and iterates through them sequentially. For this 
example, we use the table shown in Table~\ref{tab:pipeline_example}, 
which contains information about Australian license plates.

\begin{table*}[h]
\small
\resizebox{\textwidth}{!}{%
\begin{tabular}{llllll}
\toprule
\textbf{State/Territory} & \textbf{Text/bg Color} & \textbf{Format} 
& \textbf{Current Slogan} & \textbf{Current Series} & \textbf{Notes} \\
\midrule
Australian Capital Territory & blue/white & Yaa$\cdot$nna & 
ACT $\cdot$ CELEBRATION OF A CENTURY 2013 & YIL$\cdot$00A & 
Slogan screenprinted on plate \\
New South Wales & black/yellow & aa$\cdot$nn$\cdot$aa & 
NEW SOUTH WALES & BX$\cdot$99$\cdot$HI & 
No slogan on current series \\
New South Wales & black/white & aaa$\cdot$nna & NSW & CPX$\cdot$12A & 
Optional white slimline series \\
Northern Territory & ochre/white & Ca$\cdot$nn$\cdot$aa & 
NT $\cdot$ OUTBACK AUSTRALIA & CB$\cdot$06$\cdot$ZZ & 
New series began in June 2011 \\
\bottomrule
\end{tabular}%
}
\caption{Example table used in the pipeline walkthrough, drawn 
from WikiSQL. Used for demonstration purposes only.}
\label{tab:pipeline_example}
\end{table*}

\textbf{Seeding and SQL Query Generation:}
For each table, the pipeline selects one of two diversity injection 
methods with equal probability: reserved word seeding or persona seeding.

\textit{Reserved word seeding} controls SQL complexity by specifying 
a target number of SQL reserved words for the generated query. This 
count is sampled from a probability distribution derived from BIRD's 
training set, so that the synthetic data mirrors the complexity 
distribution of human-authored queries. The count is passed directly 
into the generation prompt (Prompt~\ref{fig:prompt_reserved}), and 
chain-of-thought reasoning is used to help the model adhere to the 
constraint. For example, targeting three reserved words might produce: 
\texttt{SELECT "State/Territory" FROM "table-1" WHERE 
"Text/background Color" = 'black/white'}.

\textit{Persona seeding} instead grounds generation in a specific 
user's goals and domain context. An LLM is prompted to generate a 
persona for the given table, for example, ``License plate designer 
observing trends by region,'' along with associated goals (e.g., ``To 
create the most sought-after license plate in Australia'') and example 
queries (e.g., ``Background color of Northern Territory vs. New South 
Wales''). The persona is randomly selected to be either a common 
professional type or a creative niche type to encourage diversity in 
register and terminology (Prompts~\ref{fig:prompt_persona_creative} 
and~\ref{fig:prompt_persona_common}). All three components --- persona, 
goals, and example queries --- are included in the SQL generation 
prompt (Prompt~\ref{fig:prompt_persona}).

It is important to note that reserved word seeding only influences SQL 
generation, while persona seeding influences both SQL and NL generation, 
ensuring consistency between the query and the question. All generated 
queries are executed against the database at this stage; any query that 
fails to execute is discarded.

For the remainder of this example, we assume reserved word seeding 
produced: \texttt{SELECT "State/Territory" FROM "table-1" WHERE 
"Text/background Color" = 'black/white'}, (ie, selecting states for which the background color of the license plate is black and white).

\textbf{NL Question Generation:}
The validated SQL query and table schema are used to generate a 
natural language question. When persona seeding is used, the persona 
and goals are also included, and the model is instructed to phrase the 
question in the voice of that persona, adjusting formality, 
domain-specific vocabulary, and whether the user would refer to columns 
by name or by colloquial proxy (Prompt~\ref{fig:prompt_nl_persona}). 
Without persona seeding, a standard chain-of-thought prompt is used 
(Prompt~\ref{fig:prompt_nl_gen}). In both cases, four candidate 
questions are generated simultaneously, and a compare judge selects 
the best one based on naturalness, accuracy, and completeness 
(Prompt~\ref{fig:prompt_judge}). For our example, the selected 
question might be: ``Which states use black and white as their 
license plate colors?''

\textbf{Ambiguity Detection and Handling:}
The NL question is scored by a DistilBERT-based ambiguity model 
trained on Ambrosia data \citep{ambrosia}, producing a 
continuous score from 0 to 10. This score determines how the example 
proceeds through the next stage.

\textit{If the score is 7 or above}, the question is considered 
potentially ambiguous and enters the multi-interpretation branch. A 
prompt-based filter (Prompts~\ref{fig:prompt_filter_p1} 
and~\ref{fig:prompt_filter_p2}) first confirms whether the ambiguity 
is truly present. Specifically, it tests whether it falls into one of four categories: 
Scope, Attachment, Entity Vagueness, or Broadness. The first three 
categories and their accompanying examples in the prompt are adapted 
from \citet{ambrosia}. Broadness is an additional category 
we introduce, inspired by \citet{Floratou2024NL2SQLIA}, to capture 
cases where the NL is simply underspecified (e.g., ``best player,'' 
where it is unclear whether this means most points scored or highest 
rating).  If the filter determines the question is not genuinely ambiguous, the example is discarded. If ambiguity is confirmed, the pipeline attempts to generate a second SQL query representing a distinct valid 
interpretation of the same question 
(Prompt~\ref{fig:prompt_ambiguity}). The model must also identify 
the ambiguity type and explain how the new query differs semantically 
from the original. The new query is tested for executability, and 
then passed through the same prompt filter again to verify that it 
is a genuinely different valid reinterpretation rather than a 
superficial variant. The pipeline makes up to three attempts to 
produce an acceptable alternative; if none succeed, the example 
proceeds as a standard unambiguous pair using only the original query.

For our example, ``Which states use black and white as their license 
plate colors?'' is ambiguous: ``black and white'' could mean plates 
that are both black and white together, or plates that are either 
black or white. The two valid SQL interpretations would be an 
\texttt{AND} condition and an \texttt{OR} condition respectively, 
both of which are executable and semantically distinct.

\textit{If the score is below 7}, the example skips the 
multi-interpretation branch and proceeds directly to the alignment 
model, which assigns a score between 0 and 1 reflecting how well 
the NL question and SQL query correspond to each other. Examples 
scoring 0.6 or above are retained. Note that ambiguous examples do not go through the alignment reward model, as this model was not trained to handle ambiguity. This is why they are filtered separately using prompts.

\textbf{Final Quality Filter:}
Regardless of which path an example took through the ambiguity 
stage, all examples pass through a final prompt-based filter 
(Prompt~\ref{fig:prompt_final_filter}) before being added to the 
dataset. This filter checks three things independently: that the SQL 
is a valid interpretation of the NL, that the SQL is non-redundant 
and well-formed, and that the NL sounds like something a real user 
would ask. This step catches any remaining low-quality examples that 
passed the alignment model but still contain subtle issues, such as 
alligned pairs where the question did not make sense to begin with (ex. taking the sum of a categorical column), the SQL is overly convoluted, or any other issues.

\subsection{Error Analysis}
\label{app:error_analysis}

We conducted a failure mode analysis on 100 randomly sampled 
incorrect predictions from the Qwen 2.5-3B model finetuned on our pipeline's generations, evaluated on the Spider test set. The following 
taxonomy and occurrence counts were derived from this sample.

\paragraph{Taxonomy.} We identify nine error categories:

\textbf{Acceptable variation:} The predicted SQL is semantically 
equivalent to the gold query or represents a valid alternative 
interpretation of the natural language question. This includes 
cases where an ORDER BY clause is added but not requested, an 
additional column is returned that is directly implied by the NL 
question (e.g., the question asks which player scored the most 
goals and the query returns both player name and goal count), a 
redundant but non-harmful condition is present, or minor syntactic 
differences exist that do not affect the result set. These cases 
would fail string match or execution match but have no meaningful 
impact on the user's informational need.

\textbf{Spurious predicate hallucination:} The model introduces 
a WHERE clause condition that has no basis in the natural language 
question. Example: the question asks for all employees in a 
department, but the query adds \texttt{WHERE status = `active'} 
with no such qualifier in the NL.

\textbf{Result boundary error:} The model produces the correct 
logic but returns too much or too little. For example, limiting 
to 5 results when the user implies all results, or applying 
incorrect offsets.

\textbf{Schema entity confusion:} The model maps a natural 
language entity to the wrong table or column. Example: the 
question asks for ``salary'' but the query selects ``bonus'' 
because both are numeric compensation columns.

\textbf{Logical negation failure:} The model fails to capture 
a negation expressed in the NL (e.g., ``not'', ``except'', 
``other than''), producing a query that retrieves the complement 
of the intended set. Example: ``Who did NOT win the award?'' 
returns all winners instead of all non-winners.

\textbf{Degenerate join structure:} The model produces a join 
that is structurally malformed, redundant, or semantically 
incorrect. For example, joining a table to itself unnecessarily 
or using the wrong join key.

\textbf{Cross-row constraint flattening:} Questions like ``both 
X and Y'' require set intersection across rows, but the model 
instead treats it as a condition on a single row. Example: 
``Flights that took off from Florida and New York'' becomes 
\texttt{WHERE takeoff=`Florida' AND takeoff=`New York'}, which 
can never be true on a single row.

\textbf{Hardcoded quantifier substitution:} The model sees a 
superlative such as ``highest'' or ``fewest'' and instead of 
computing the answer from the data, substitutes a hardcoded 
number. Example: ``rated higher than average'' becomes 
\texttt{WHERE rating > 4}, which will be wrong on any dataset 
where the true average differs.

\textbf{NL misinterpretation:} The model misreads the core 
intent of the NL question and produces a query that answers 
a related but distinct question. Example: ``How many orders 
were placed last month?'' is answered with a list of all orders 
rather than a count.

\subsection{Complexity and Performance}
\label{app:complex}

Table~\ref{tab:llm_experiments} demonstrates our pipeline's model agnosticism: generating data with models ranging from Qwen-2.5-3B to Llama-3.3-70B produces similar SQL complexity and NL richness. Filtering leaves SQL complexity largely stable while modestly improving NL richness. Notably, the alignment reward model was trained on BIRD-derived pairs, yet filtering does not systematically shift complexity toward BIRD's distribution — this confirms the filter selects for NL-SQL coherence rather than acting as a proxy for BIRD-like complexity, confirming that using BIRD as both the target distribution and the source of alignment training data does not introduce circularity.

\begin{table}[h!]
  \centering
  \begin{tabular}{lcc}
    \hline
    \textbf{Model} & \textbf{SQL Comp.} & \textbf{NL Rich.} \\
    \hline
    NF Qwen-3B & 0.38 & 0.40 \\
    NF Llama-8B & 0.43 &  0.37 \\
    NF Qwen-32B & 0.37 & 0.37 \\
    NF Llama-70B & 0.41 & 0.40 \\
    \hline
    F Qwen-3B & 0.38 & 0.41 \\
    F Llama-8B & 0.39 &  0.43\\
    F Qwen-32B & 0.38 & 0.40\\
    F Llama-70B & 0.45 & 0.39\\
    \hline    
  \end{tabular}
  \caption{SQL complexity and NL richness across different models (seeded with WikiSQL) before and after filtering, showing no significant differences between complexity. Before filtering is indicated by NF (No-Filtering) and after filtering is indicated by F (Filtering). All metrics were calculated on a sample size of 100 for each model. We used the following models: Qwen-2.5-3B, Llama-3.1-8B, Qwen-3-32B, and Llama-3.3-70B.}
  \label{tab:llm_experiments}
\end{table}

\subsection{Cost Analysis}
\label{app:complex}

\label{app:cost}

The cost estimates rest on several assumptions. 
For token counts, we estimate an average of 5 LLM calls per example based on 
the mandatory pipeline stages: SQL generation, NL generation, compare judge, 
and final prompt filter. Persona generation adds one additional call for 
approximately 50\% of examples; we treat 5 as a conservative average across 
both persona-seeded and reserved-word-seeded examples. Input and output token 
counts were estimated by populating prompt templates with representative 
examples sampled from the middle of our output log distribution, measured 
using the GPT-5 tokenizer, yielding 5,265 input tokens and 784 output tokens 
per example. Based on our reward model statistics and scaled quality 
measurements across models, we estimate a retention rate of 84\%, meaning 
6,548 examples must be generated to yield 5,500 kept examples. This gives a 
total of approximately 34.5 million input tokens and 5.1 million output tokens. 
At GPT-5.5 pricing (\$5.00/1M input tokens, \$30.00/1M output tokens), an 
equivalent API-based run would cost approximately \$326. Local runtime was 
estimated assuming input prefill at approximately 2,000 tokens per second and 
output generation at approximately 50--80 tokens per second on a single 
commodity GPU running Qwen-2.5-3B, yielding approximately 136 GPU-hours and 
a cost of approximately \$20 at \$0.15/hr spot pricing (e.g., Vast.ai). This 
represents a 16$\times$ cost reduction, before accounting for the privacy 
implications of transmitting proprietary schemas to third-party APIs. The 
alignment reward model and DistilBERT ambiguity scorer are local model 
inference and contribute negligible additional cost. All API pricing reflects 
publicly listed rates at time of submission, and GPT 5.5 was chosen due to its status as a popular proprietary model at time of submission.

\subsection{SQL Complexity Distribution Analysis}
\label{app:complexity_distribution}

\begin{figure}[h]
    \centering
    \includegraphics[width=1\linewidth]{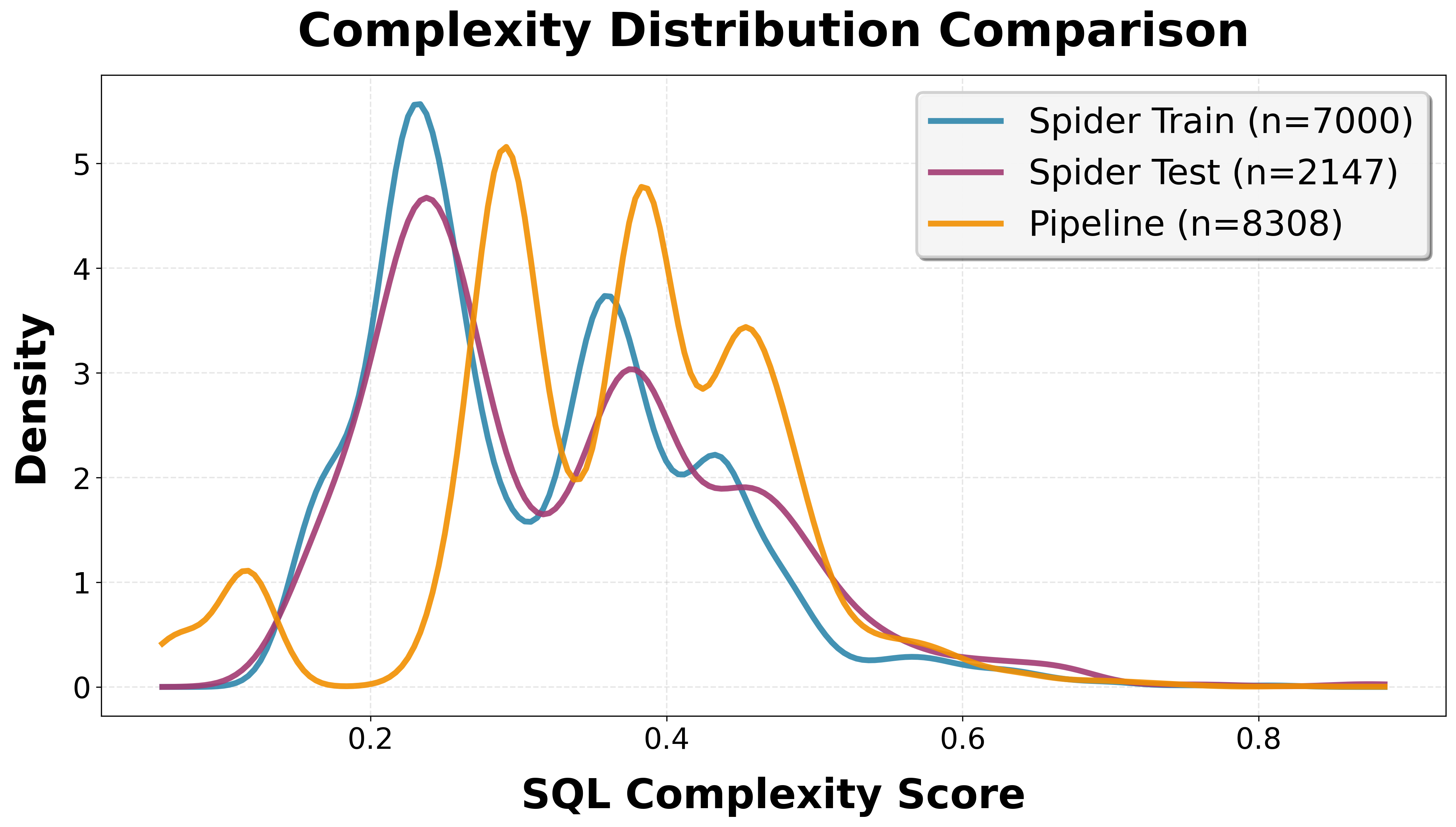}
    \caption{SQL complexity distributions for Spider train (n=7,000), Spider test (n=2,147), and FALCON-generated data (n=8,308). FALCON data exhibits higher complexity with bimodal peaks and extends substantially into harder query ranges (0.4-0.8) underrepresented in Spider.}
    \label{fig:complexity_dist}
\end{figure}

Figure~\ref{fig:complexity_dist} shows the SQL complexity distributions across datasets. Spider train and test sets exhibit nearly identical distributions (blue and purple overlapping curves), both centered around 0.2 complexity with minimal representation beyond 0.4. In contrast, our FALCON-generated data (orange) shows a distinct rightward shift with bimodal peaks at 0.35 and 0.45, and substantial density extending beyond 0.5. This distributional difference explains the cross-dataset performance gap in Table~\ref{tab:cross_dataset_eval_3b}: Spider-trained models struggle on FALCON data because they lack exposure to queries in the 0.4-0.6 complexity range that dominate our synthetic dataset, while FALCON-trained models do not generalize as well to easier queries.

\subsection{Alignment Model}
\label{app:alignment}

\paragraph{Training Data Creation.} We create a supervised dataset from BIRD SQL queries by generating candidate NL and CoT pairs for each query using Qwen-2.5-3B. To prevent overfitting to perfect alignments, we corrupt 50\% of examples with 1-5 modifications per query (e.g., altering \texttt{WHERE} clauses, etc.) The resulting dataset contains over 9,000 examples (median alignment: 0.72, IQR: 0.54-0.90).
\paragraph{Alignment Score Computation.} For each example, we compute a hybrid alignment score $y \in [0, 1]$ by comparing the generated question to BIRD's human-written reference question. Because BIRD questions align with their SQL queries, similarity to the reference indicates how well the generated question aligns with the SQL. Traditional semantic similarity (e.g., Sentence-BERT) often misses critical differences; For instance, "return the average" and "return the highest" may score similarly despite requiring different aggregations (AVG vs. MAX).
Our metric combines two weighted components:
\begin{enumerate}
    \item \textbf{Semantic Similarity (40\%):} Sentence-BERT embeddings measuring linguistic proximity.
    \item \textbf{Schema Entity Scoring (60\%):} Extracts entities (table/column names, literal values) from the SQL query and detects their mentions in the NL question. Applies multiplicative penalties ($0.7^n$) for missing ground-truth entities or incorrect entities.
\end{enumerate}
For corrupted examples, we apply an additional 10\% penalty per corruption (Figure~\ref{fig:corruption_similarity_example}). We validate our computed alignment scores against human annotations on 101 examples, achieving strong correlation (Pearson correlation $r=0.70$; Appendix~\ref{tab:alignment_correlation}). We train ModernBERT on this dataset, achieving MSE: 0.0203 and MAE: 0.1129 (Table~\ref{tab:alignment_models}). 
This alignment model outperforms other filtering methods when applied to our main synthetic dataset (Table~\ref{tab:quality_filtering}).
\begin{figure}[ht]
\centering
\begin{tcolorbox}[
    colback=filteringorange!20, 
    colframe=black!50, 
    boxrule=0.5pt,
    boxsep=3pt,
    left=5pt,
    right=5pt,
    top=5pt,
    bottom=5pt
]
\textbf{Predicted NL:} What unique user ID rates 'The Matrix' with a score of 1?''
\textbf{True NL:} List the users who gave the worst rating for movie 'Love Will Tear Us Apart'.''
\vspace{3pt}
\hrule
\vspace{3pt}
\textbf{Similarity Computation:}
\textbf{1. Sentence-BERT Similarity:} 0.67
\vspace{3pt}
\textbf{2. Entity-Based Score:} 0.49
Missing true entity: Love Will Tear Us Apart'' (penalty: $0.7^1 = 0.70$)
Corrupted entity mentioned: The Matrix'' (penalty: $0.7^1 = 0.70$)
Entity score: $1.0 \times 0.70 \times 0.70 = 0.49$
\vspace{3pt}
\textbf{3. Weighted Similarity:}
\[
0.67 \times 0.4 + 0.49 \times 0.6 = 0.268 + 0.294 = 0.562
\]
\textbf{4. Corruption Penalty:} $0.562 \times 0.9 = \mathbf{0.506}$
\end{tcolorbox}
\caption{Similarity scoring example. Despite high sentence similarity (0.67), entity penalties and corruption penalty reduce the final score to 0.506.}
\label{fig:corruption_similarity_example}
\end{figure}

Human-annotators evaluated 101 examples using our rubric (See~\ref{app:rubric} for rubric), mapping scores of 1, 2, and 3 to 0, 0.5, and 1 respectively. Our computed alignment scores using this metric achieved a Pearson correlation of 0.70 with human judgments. Some variance is expected when comparing our continuous scores (0-1) to discrete three-interval annotations.
\begin{table}[h!]
\centering
\begin{tabular}{lc}
\toprule
\textbf{Metric} & \textbf{Value} \\
\midrule
\multicolumn{2}{l}{\textit{Correlation Metrics}} \\
Pearson's r & 0.701 \\
Spearman's $\rho$ & 0.700 \\
\bottomrule
\end{tabular}
\caption{Correlation between human evaluation scores and alignment metric on 101 SQL-to-NL pairs ($p < 0.001$). Human evaluators scored examples as 0, 0.5, or 1.0 according to our rubric (Table~\ref{app:rubric}). Alignment scores were computed using our hybrid similarity metric combining semantic and schema-entity matching.}
\label{tab:alignment_correlation}
\end{table}

We trained three model architectures on this dataset: RLVR with Qwen, regression-based CodeBERT, and regression-based ModernBERT. ModernBERT achieved the lowest error rates (MSE: 0.0203, MAE: 0.1129) with compact size (Table~\ref{tab:alignment_models}). We trained this model with learning rate $2e^{-5}$, batch size 64, and early stopping (patience: 5). 
\begin{table}[h!]
\centering
    \begin{tabular}{lcc}
    \hline
    \textbf{Model} & \textbf{MSE} & \textbf{MAE}\\
    \hline
    RLVR w/ Qwen & 0.0341 & 0.1506\\
    FT + CodeBERT & 0.0268 & 0.1262 \\
    FT + ModernBERT & \textbf{0.0203} & \textbf{0.1129} \\
    \hline
  \end{tabular}
  \caption{Alignment model performance on held-out test set.}
  \label{tab:alignment_models}
\end{table}

\subsection{Human Evaluation Across Seed Databases}
We compare human evaluation across seed databases, showing that high human quality scores hold. Four annotators independently scored disjoint subsets of 100 examples each, with disagreements resolved through group discussion.
\begin{table}[t]
  \centering
  \begin{tabular}{lcccc}
    \hline
    \textbf{Model} & \textbf{SQL} & \textbf{NL} & \textbf{Alignment} \\
    \hline
    WikiSQL  & 0.92 & 0.94 & 0.95 \\
    Spider  & 0.99 & 0.95 & 0.96 \\
    \hline    
  \end{tabular}
  \caption{Human evaluation scores across databases on Llama-3.1-70B.}
  \label{tab:human_eval_cross_databases}
\end{table}

\subsection{Ambiguity Model Training.} 
\label{app:ambiguity_model_training}

The ambiguity model is trained on Ambrosia-style data \citep{ambrosia}, where each question has (i) a canonical SQL query \texttt{gold\_query}, (ii) additional valid interpretations \texttt{ambig\_queries}, and (iii) a boolean \texttt{is\_ambiguous} flag. We count distinct interpretations as $k_i = \left|\{\texttt{gold\_query}\} \cup \texttt{ambig\_queries}\right|$. This count maps to a score $y_i \in [0, 10]$: unambiguous questions receive $y_i = 0$, while ambiguous questions are mapped from $k_i$ to $[5, 10]$ via min-max normalization. This ensures scores reflect interpretation count rather than schema-specific artifacts.

\paragraph{Pipeline Integration.}
Predicted scores $\hat{y}$ influence two pipeline stages: (1) \textbf{Filtering} removes questions with extreme ambiguity scores (representing very unclear, low quality NL questions), and (2) \textbf{Multi-Interpretation Generation} triggers when $\hat{y} > 7$, generating a second plausible SQL query to mirror Ambrosia's structure and provide richer supervision for ambiguous questions.

\subsection{Annotation Rubric}
\label{app:rubric}

Found at Table~\ref{tab:rubric}.

\begin{table*}[h]
\centering
\begin{tabular}{p{0.15\textwidth}cp{0.65\textwidth}}
\toprule
\textbf{Metric} & \textbf{Score} & \textbf{Description} \\
\midrule
\multirow{3}{*}{\textbf{SQL-NL Alignment}} & 1.0 & \textbf{Correct Match:} Covers all NL requirements. Minor relevant additions (e.g., adding a "points" column when asking for "who scored most", setting a limit value, etc.) are permitted. \\
\cmidrule{2-3}
& 0.5 & \textbf{Minor Issues:} Mostly correct but adds unrelated details (e.g., groups by category when not asked to) or misses 1-2 small parts of a complex request. \\
\cmidrule{2-3}
& 0.0 & \textbf{Incorrect:} Major omissions or fundamental misunderstanding of the NL intent. \\
\midrule
\multirow{3}{*}{\textbf{SQL Quality}} & 1.0 & \textbf{Good:} Efficient structure and standard formatting. \\
\cmidrule{2-3}
& 0.5 & \textbf{Functional:} Query works but is slightly inefficient or unnecessarily complex. \\
\cmidrule{2-3}
& 0.0 & \textbf{Poor:} Query is nonsense, invalid, or extremely convoluted. \\
\midrule
\multirow{3}{*}{\textbf{NL Quality}} & 1.0 & \textbf{Natural:} The question is sensical and sounds like human speech. \\
\cmidrule{2-3}
& 0.5 & \textbf{Awkward:} Understandable intent, but phrased in an unnatural or robotic way. \\
\cmidrule{2-3}
& 0.0 & \textbf{Nonsensical:} Logic errors (e.g., asking for the average of a categorical variable) or incoherent phrasing. \\
\bottomrule
\end{tabular}
\caption{The three-point scoring rubric used for human evaluation of synthetic NL-SQL pairs.}
\label{tab:rubric}
\end{table*}

\subsection{Examples from the Dataset}
\label{app:examples}

We provide four representative examples from the FALCON-generated 
dataset to illustrate the range of query styles and complexity levels 
produced by the pipeline. (Figure~\ref{fig:example1}, Figure~\ref{fig:example2}, Figure~\ref{fig:example3}, Figure~\ref{fig:example4})

\begin{figure*}[t]
\begin{mdframed}
\noindent\textbf{Example 1: Casual query.}

\smallskip
\noindent\textit{Persona (shortened for brevity):} Amanda Thompson is a travel 
blogger who uses storm record data to create safety guides for her readers.

\smallskip
\noindent\textit{NL:} ``What are the 5 regions that have experienced the most 
storms in the past 10 years, and what are their region names and codes, 
so I can warn my readers about these storm-prone areas?''

\begin{lstlisting}
SELECT `region`.`region name`, `region`.`region code`,
       COUNT(`storm`.`storm id`) AS `storm frequency`
FROM `region`
JOIN `affected region`
     ON `region`.`region id` = `affected region`.`region id`
JOIN `storm`
     ON `affected region`.`storm id` = `storm`.`storm id`
WHERE `storm`.`storm date` > DATE('now', '-10 years')
GROUP BY `region`.`region name`, `region`.`region code`
ORDER BY `storm frequency` DESC LIMIT 5;
\end{lstlisting}

\noindent This example illustrates how persona seeding produces natural, 
conversational queries grounded in a believable real-world motivation. 
The trailing context (``so I can warn my readers'') reflects how actual 
users frame database questions around their goals rather than as 
abstract data retrieval tasks, and the SQL faithfully captures the 
intent including the temporal filter and ranking.
\end{mdframed}
\caption{Example 1: Casual query generated via persona seeding.}
\label{fig:example1}
\end{figure*}

\begin{figure*}[t]
\begin{mdframed}
\noindent\textbf{Example 2: Formal query.}

\smallskip
\noindent\textit{Persona (shortened for brevity):} Emilia Rodriguez is a senior 
real estate analyst at a property investment firm, responsible for 
identifying profitable investment opportunities.

\smallskip
\noindent\textit{NL:} ``Can I get a detailed report showing the property type 
descriptions, their average agreed selling prices, and the number of 
properties for each type, all filtered for properties with more than 
5 rooms and selling prices over \$1 million, and sorted by the average 
selling price in descending order?''

\begin{lstlisting}
SELECT `rpt`.`property type description`,
       AVG(`p`.`agreed selling price`) AS `average_agreed_selling_price`,
       COUNT(`p`.`property id`) AS `number_of_properties`
FROM `properties` `p`
JOIN `reference property types` `rpt`
     ON `p`.`property type code` = `rpt`.`property type code`
WHERE `p`.`room count` > 5
  AND `p`.`agreed selling price` > 1000000
GROUP BY `rpt`.`property type description`
ORDER BY `average_agreed_selling_price` DESC;
\end{lstlisting}

\noindent This example demonstrates that the pipeline also produces formal, 
technically precise queries with complex filtering and aggregation. 
The persona provides a credible motivation for the specificity of 
the query: an analyst seeking to identify the highest-value property 
types within the premium market segment, explaining why the query 
scopes to large, high-price properties rather than the full dataset.
\end{mdframed}
\caption{Example 2: Formal query generated via persona seeding.}
\label{fig:example2}
\end{figure*}

\begin{figure*}[t]
\begin{mdframed}
\noindent\textbf{Example 3: Ambiguous query with multiple valid interpretations.}

\smallskip
\noindent\textit{NL:} ``How can I obtain a list of patients who had procedures 
costing more than \$5,000, including their names, addresses, and 
hospital stay details, as well as the names of the physicians and 
nurses who cared for them?''

\smallskip
\noindent\textit{Interpretation 1} (per-procedure filter):

\begin{lstlisting}
SELECT `patient`.`name`, `patient`.`address`,
       `stay`.`stay start`, `stay`.`stay end`,
       `procedures`.`name` AS `procedure_name`,
       `physician`.`name` AS `physician_name`,
       `nurse`.`name` AS `nurse_name`
FROM `patient`
JOIN `undergoes` ON `patient`.`ssn` = `undergoes`.`patient`
JOIN `stay` ON `undergoes`.`stay` = `stay`.`stay id`
JOIN `procedures` ON `undergoes`.`procedures` = `procedures`.`code`
JOIN `physician` ON `undergoes`.`physician` = `physician`.`employee id`
JOIN `nurse` ON `undergoes`.`assisting nurse` = `nurse`.`employee id`
WHERE `procedures`.`cost` > 5000.0
ORDER BY `stay`.`stay end` - `stay`.`stay start` DESC;
\end{lstlisting}

\smallskip
\noindent\textit{Interpretation 2} (total cost filter):

\begin{lstlisting}
SELECT `patient`.`name`, `patient`.`address`,
       `stay`.`stay start`, `stay`.`stay end`,
       `physician`.`name` AS `physician_name`,
       `nurse`.`name` AS `nurse_name`
FROM `patient`
JOIN `undergoes` ON `patient`.`ssn` = `undergoes`.`patient`
JOIN `stay` ON `undergoes`.`stay` = `stay`.`stay id`
JOIN `procedures` ON `undergoes`.`procedures` = `procedures`.`code`
JOIN `physician` ON `undergoes`.`physician` = `physician`.`employee id`
JOIN `nurse` ON `undergoes`.`assisting nurse` = `nurse`.`employee id`
WHERE `undergoes`.`patient` IN (
    SELECT `undergoes`.`patient` FROM `undergoes`
    JOIN `procedures`
         ON `undergoes`.`procedures` = `procedures`.`code`
    GROUP BY `undergoes`.`patient`
    HAVING SUM(`procedures`.`cost`) > 5000.0
)
ORDER BY `stay`.`stay end` - `stay`.`stay start` DESC;
\end{lstlisting}

\noindent The phrase ``procedures costing more than \$5,000'' admits two valid 
interpretations: a per-procedure filter (any single procedure exceeds 
\$5,000) or a total cost filter (the sum of all procedures for a 
patient exceeds \$5,000). The pipeline detects this as a Broadness 
ambiguity and generates both SQL interpretations as valid supervision 
signal.
\end{mdframed}
\caption{Example 3: Ambiguous query with two valid SQL interpretations.}
\label{fig:example3}
\end{figure*}

\begin{figure*}[t]
\begin{mdframed}
\noindent\textbf{Example 4: Reserved word seeding.}

\smallskip
\noindent\textit{NL:} ``In 2016, which events were held in stadiums that can 
accommodate more than 60,000 people, listed in alphabetical order by 
stadium name?''

\begin{lstlisting}
SELECT `stadium`.`name`, `event`.`name`
FROM `stadium`
JOIN `event` ON `stadium`.`id` = `event`.`stadium id`
WHERE `event`.`year` = '2016'
  AND `stadium`.`capacity` > 60000
ORDER BY `stadium`.`name` ASC;
\end{lstlisting}

\noindent This example illustrates how reserved word seeding controls query 
complexity by grounding it in the distribution of reserved word counts 
observed in BIRD. The seeding value of 9 is sampled from this 
distribution, and the model is prompted to generate a query containing 
that many reserved words. The resulting SQL faithfully reflects the 
target complexity, joining two tables on a shared identifier and 
applying both a temporal filter and a capacity threshold before 
ordering results alphabetically by stadium name.
\end{mdframed}
\caption{Example 4: Query generated via reserved word seeding.}
\label{fig:example4}
\end{figure*}

\subsection{Qwen 2.5-3B Finetuning and Results}
\label{app:qwen3btune}
Models are fine-tuned using LoRA (rank=16, $\alpha$=32, dropout=0.05) for 3 epochs with learning rate $2\times10^{-4}$, effective batch size 16 (via gradient accumulation), and completion-only masking that trains solely on SQL outputs while masking instructions and schema. Training uses up to 5,500 examples sampled randomly when datasets exceed this size. Table schemas provided to the model include column names, types, primary keys, and foreign key relationships to connected tables. Models are evaluated on both our held-out synthetic test set (n=2,000) \textbf{and the Spider test set (n=2,147), a real human-authored benchmark entirely independent of our generation pipeline}. We do not evaluate on BIRD's test set because our alignment reward model was trained on BIRD-derived pairs, which would confound evaluation; Spider serves as our independent real-world generalization benchmark. All three training conditions — synthetic-only, Spider-only, and mixed — are evaluated on both test sets to enable direct cross-dataset comparison.

\subsection{Qwen 2.5-1.5B Training and Results}
\label{app:qwen1btune}
We supplement our downstream task performance with results from a smaller model (Qwen 2.5-1.5B). Models are fine-tuned using LoRA (rank=16, $\alpha$=32) for 2 epochs with learning rate $2\times10^{-4}$, effective batch size 16 (via gradient accumulation), and completion-only masking that trains solely on SQL outputs while masking instructions and schema. Training uses 5,500 examples with 2,000 held-out for testing. Due to the more compact model size, only relevant tables appearing in the gold query are included in the table schema. Models are fine-tuned for 2 epochs; loss continued to decrease at the end of training, suggesting additional epochs may further 
improve performance.
\label{app:qwen_variants}
\begin{table*}[h]
\centering
\begin{tabular}{lccccccc}
\toprule
\textbf{Model} & \textbf{Test Set} & \textbf{Total} & \textbf{EM \%} & \textbf{FM \%} & \textbf{FS} & \textbf{Exec \%} \\
\midrule
Spider (5.5k) & Spider & 2147 & 31.21 & 52.77 & 0.87 & 49.65 \\
Spider (5.5k) & Pipeline & 2000 & 0.1 & 3.2 & 0.66 & 26.72 \\
Pipeline (5.5k) & Spider & 2147 & 1.07 & 8.52 & 0.70 & 72.71 \\
Pipeline (5.5k) & Pipeline & 2000 & 9.1 & 42.55 & 0.83 & 84.11 \\
Mixed 5\% (5.5k) & Spider & 2147 & 22.82 & 38.94 & 0.81 & -- \\
Mixed 5\% (5.5k) & Pipeline & 2000 & 9.2 & 42.8 & 0.83 & -- \\
Mixed 10\% (5.5k) & Spider & 2147 & 27.53 & 44.67 & 0.84 & -- \\
Mixed 10\% (5.5k) & Pipeline & 2000 & 9.65 & 44.05 & 0.83 & -- \\
\bottomrule
\end{tabular}
\caption{Performance of Qwen 2.5-1.5B LoRA variants (2 epochs) on Spider and FALCON test sets. Training data sizes shown in parentheses. \textbf{Metrics:} EM = Exact Match (string equivalence), FM = Fuzzy Match at 90\%, FS = Fuzzy Score (average string similarity), Exec = Execution Accuracy (result set overlap). \textbf{Training data:} Spider = real Spider data, Pipeline = FALCON-generated data, Mixed = combined data (5\% or 10\% real Spider + remaining pipeline). Models trained on FALCON data excel on FALCON tests, while Spider-trained models excel on Spider tests, demonstrating training distribution effects.}
\label{tab:qwen_variants}
\end{table*}

\begin{figure*}[t]
\begin{tcolorbox}[
    colback=gray!5!white,
    colframe=gray!75!black,
    title=SQL Generation with Reserved Word Seeding,
    fontupper=\small\ttfamily,
    width=\textwidth
]
You are an expert SQL developer tasked with generating a creative\\
and non-trivial SQLite query.\\
\\
<database\_schema>\\
\{table\}\\
</database\_schema>\\
\\
<task\_requirements>\\
1. Source Material: Construct the query using ONLY information\\
~~~from the <database\_schema> provided above.\\
2. Logic: The query must be executable, logically sound, and\\
~~~produce actionable insights.\\
3. Identifier Formatting: You must encase ALL column names in\\
~~~backticks (`) and all string literals in single quotes.\\
4. Formatting: Output a single SQLite query inside ```sql\\
~~~code blocks.\\
5. Syntax: End with a semicolon (;).\\
</task\_requirements>\\
\\
<constraint\_checklist>\\
- The query must contain EXACTLY \{length\_seed\} reserved words.\\
- Reserved words to choose from: \{reserved\_word\_patterns\}\\
- Do not use fewer or more than \{length\_seed\}.\\
</constraint\_checklist>\\
\\
Step-by-step instructions:\\
1. Analyze the table schema.\\
2. Use exactly \{length\_seed\} reserved words from the list.\\
3. Construct a query that uses exactly those reserved words\\
~~~and answers a non-trivial question.\\
4. Verify backtick formatting.\\
\\
Output ONLY the SQL query inside code blocks. No explanations.
\end{tcolorbox}
\caption{SQL generation prompt with reserved-word seeding.}
\label{fig:prompt_reserved}
\end{figure*}

\begin{figure*}[t]
\begin{tcolorbox}[
    colback=gray!5!white,
    colframe=gray!75!black,
    title=SQL Generation with Persona Seeding,
    fontupper=\small\ttfamily,
    width=\textwidth
]
You are an expert SQL developer. Generate a SQLite query that\\
matches the specific goals of the user persona described below.\\
\\
<database\_schema>\\
\{table\}\\
</database\_schema>\\
\\
<persona\_profile>\\
**Persona:** \{persona\}\\
**Goals:** \{goals\}\\
**Inspiration Queries:** \{example\_queries\}\\
</persona\_profile>\\
\\
<instructions>\\
1. Relevance: The query must answer a question this specific\\
~~~persona would genuinely ask. Use the inspiration queries\\
~~~as a guide for complexity, but do not copy them exactly.\\
2. Strict Formatting:\\
~~~- Use valid SQLite syntax.\\
~~~- Encase ALL column names in backticks (e.g., `Column Name`).\\
~~~- All string literals in single quotes (e.g. 'New South Wales').\\
3. Output: A single executable SQL query inside ```sql tags,\\
~~~ending with a semicolon.\\
4. Efficiency: Minimal steps to achieve the insight.\\
</instructions>\\
\\
Output ONLY the SQL query. No natural language.
\end{tcolorbox}
\caption{SQL generation prompt with persona seeding.}
\label{fig:prompt_persona}
\end{figure*}

\begin{figure*}[t]
\begin{tcolorbox}[
    colback=gray!5!white,
    colframe=gray!75!black,
    title=Out-of-the-Box Persona Generation,
    fontupper=\small\ttfamily,
    width=\textwidth
]
You are tasked with inventing a creative yet plausible persona\\
who interprets the provided database. This should be a person\\
who does not have a background in SQL or data analysis, but\\
wants to use the data for a relevant hobby, question, or other\\
job related use case.\\
\\
<database\_schema>\\
\{table\}\\
</database\_schema>\\
\\
<instructions>\\
Create a single persona. Their motivation must be grounded in\\
the data provided.\\
\\
Output your response as a JSON object following this exact schema:\\
\{\\
~~~~"persona\_description": "Who they are and what makes them\\
~~~~~~~~unique (Max 1 sentence).",\\
~~~~"goals": "What they are trying to achieve (Imaginative\\
~~~~~~~~but plausible).",\\
~~~~"example\_queries": [\\
~~~~~~~~"Query idea 1 (Natural language or SQL description)",\\
~~~~~~~~"Query idea 2",\\
~~~~~~~~"Query idea 3"\\
~~~~]\\
\}\\
</instructions>
\end{tcolorbox}
\caption{Creative persona generation prompt.}
\label{fig:prompt_persona_creative}
\end{figure*}

\begin{figure*}[t]
\begin{tcolorbox}[
    colback=gray!5!white,
    colframe=gray!75!black,
    title=Common Persona Generation,
    fontupper=\small\ttfamily,
    width=\textwidth
]
You are tasked with generating a realistic and likely persona\\
who would query this database in a professional capacity.\\
\\
<database\_schema>\\
\{table\}\\
</database\_schema>\\
\\
<instructions>\\
Create a single, common persona. Avoid generic or previously\\
used personas.\\
\\
Output your response as a JSON object following this exact schema:\\
\{\\
~~~~"persona\_description": "Role and background (Max 2 sentences).",\\
~~~~"goals": "What they are trying to analyze.",\\
~~~~"example\_queries": [\\
~~~~~~~~"Practical query idea 1",\\
~~~~~~~~"Practical query idea 2",\\
~~~~~~~~"Practical query idea 3"\\
~~~~]\\
\}\\
</instructions>
\end{tcolorbox}
\caption{Common persona generation prompt.}
\label{fig:prompt_persona_common}
\end{figure*}

\begin{figure*}[t]
\begin{tcolorbox}[
    colback=gray!5!white,
    colframe=gray!75!black,
    title=Natural Language Question Generation,
    fontupper=\small\ttfamily,
    width=\textwidth
]
You are a natural language expert. Translate the SQLite query\\
below into 4 distinct natural language questions.\\
\\
<context>\\
<database\_schema>\\
\{table\}\\
</database\_schema>\\
\\
<sql\_query>\\
\{sql\_query\}\\
</sql\_query>\\
</context>\\
\\
<instructions>\\
Analyze the SQL intent and generate 4 questions that logically\\
lead to this query.\\
1. No SQL Leakage: Do not use technical terms (SELECT, GROUP BY)\\
~~~or column names with underscores. Use natural English.\\
2. Accuracy: Do not infer details not present in the query.\\
~~~Any implications should be clear and specified.\\
3. Variety: Phrase the questions differently (e.g., different\\
~~~levels of formality or sentence structure).\\
\\
Structure your thinking for EACH question:\\
1. Parse SQL Intent -> 2. Map to Schema -> 3. Draft Question\\
-> 4. Validate.\\
\\
Output must be a single JSON object with this structure:\\
\{\\
~~~~"reasoning1": "Analysis for Q1 (Max 4 sentences)",\\
~~~~"question1": "Natural language question 1",\\
~~~~...\\
\}\\
</instructions>
\end{tcolorbox}
\caption{NL question generation prompt with chain-of-thought.}
\label{fig:prompt_nl_gen}
\end{figure*}

\begin{figure*}[t]
\begin{tcolorbox}[
    colback=gray!5!white,
    colframe=gray!75!black,
    title=Natural Language Question Generation with Persona Voice,
    fontupper=\small\ttfamily,
    width=\textwidth
]
You are a natural language expert. Translate the SQLite query\\
into 4 questions using the voice and style of a specific persona.\\
\\
<context>\\
<database\_schema>\\
\{table\}\\
</database\_schema>\\
\\
<sql\_query>\\
\{sql\_query\}\\
</sql\_query>\\
\\
<persona\_profile>\\
**Persona:** \{persona\}\\
**Goals:** \{goals\}\\
</persona\_profile>\\
</context>\\
\\
<instructions>\\
Generate 4 natural language questions that this specific persona\\
would ask to get this data.\\
1. **Voice \& Tone:** Mimic their likely jargon, casualness,\\
~~~or strictness.\\
~~~- Does this persona know SQL terms? Or do they use vague\\
~~~~~business terms?\\
~~~- Are they polite or demanding?\\
2. **Accuracy:** The question must be answerable by the provided\\
~~~SQL. Any implications should be clear and specified. For\\
~~~example, if the SQL filters on over 100 people, the question\\
~~~should say something like "big rooms (>100 people)" or\\
~~~something similar to make the implication clear.\\
\\
Output must be a single JSON object:\\
\{\\
~~~~"reasoning1": "How the persona views this data (Max 2 sentences)",\\
~~~~"question1": "The question in the persona's voice",\\
~~~~"reasoning2": "...",\\
~~~~"question2": "...",\\
~~~~"reasoning3": "...",\\
~~~~"question3": "...",\\
~~~~"reasoning4": "...",\\
~~~~"question4": "..."\\
\}\\
</instructions>
\end{tcolorbox}
\caption{NL question generation with persona-specific voice and tone.}
\label{fig:prompt_nl_persona}
\end{figure*}

\begin{figure*}[t]
\begin{tcolorbox}[
    colback=gray!5!white,
    colframe=gray!75!black,
    title=Compare Judge for NL Candidate Selection,
    fontupper=\small\ttfamily,
    width=\textwidth
]
You are an expert evaluator of Natural Language to SQL datasets.\\
\\
<task>\\
Select the best Natural Language Question from the options provided.\\
</task>\\
\\
<inputs>\\
<database\_schema>\\
\{table\}\\
</database\_schema>\\
\\
<sql\_query>\\
\{sql\_query\}\\
</sql\_query>\\
\\
<candidate\_questions>\\
\{nl\_question\}\\
</candidate\_questions>\\
</inputs>\\
\\
<evaluation\_criteria>\\
1. Completeness: The question implies all filters/logic\\
~~~present in the SQL.\\
2. No Hallucination: It does not ask for data absent from\\
~~~the SQL.\\
3. Naturalness: It sounds like a human wrote it (no underscores,\\
~~~no "SELECT *").\\
4. Independence: The question must make sense without seeing\\
~~~the table schema.\\
</evaluation\_criteria>\\
\\
Output as JSON:\\
\{\\
~~~~"index": <integer\_index\_of\_best\_question>,\\
~~~~"question": "<text\_of\_best\_question>"\\
\}
\end{tcolorbox}
\caption{LLM-as-a-judge prompt for NL candidate selection. (Compare)}
\label{fig:prompt_judge}
\end{figure*}

\begin{figure*}[t]
\begin{tcolorbox}[
    colback=gray!5!white,
    colframe=gray!75!black,
    title=Score Judge for NL Candidate Selection,
    fontupper=\small\ttfamily,
    width=\textwidth
]
You are an expert natural language expert tasked with correcting a natural language question.\\
Check whether or not the natural language question is valid based on the SQLite query and table information and provide a score between 0 and 100.\\
\\
Your score should be based on the following criteria:\\
1. The natural language question should produce a question. It does not include the SQLite query or table information.\\
2. The natural language question should be specific enough to be answered by the SQLite query.\\
3. The natural language question should not include details that the SQLite query does not contain.\\
4. The natural language question should not include any information besides the question.\\
5. The natural language question should only be one singular natural language question.\\
\\
Natural Language Question: \{nl\_question\}\\
\\
SQLite Query: \{sql\_query\}\\
\\
 Table Information: \{table\}\\
\\
Format your response as a JSON object with the following keys:\\
- "score": the score between 0 and 100\\
- "reasoning": the reasoning for the score\\
\\
Example response:\\
```json\\
\{\{\\
    "score": 80,\\
    "reasoning": "The natural language question is valid based on the SQLite query and table information but isn't specific enough to be answered by the SQLite query."\\
    \}\}\\
```\\
\\
Do not include any other text in your response besides the JSON object.

\end{tcolorbox}
\caption{LLM-as-a-judge prompt for NL candidate selection. (Score)}
\label{fig:score_judge}
\end{figure*}

\begin{figure*}[t]
\begin{tcolorbox}[
    colback=gray!5!white,
    colframe=gray!75!black,
    title=Ambiguity Prompt-Based Filter (Part 1),
    fontupper=\small\ttfamily,
    width=\textwidth
]
You are an expert in SQL interpretation and ambiguity detection.\\
<task>\\
Determine if the provided SQL query is a valid interpretation\\
of the potentially ambiguous Natural Language (NL) question.\\
</task>\\
\\
<inputs>\\
<nl\_question>\\
\{nl\_question\}\\
</nl\_question>\\
<sql\_query>\\
\{sql\_query\}\\
</sql\_query>\\
<database\_schema>\\
\{table\}\\
</database\_schema>\\
<ambiguity\_reasoning>\\
\{reasoning\}\\
</ambiguity\_reasoning>\\
</inputs>\\
\\
<ambiguity\_types>\\
Only the following ambiguity types are considered valid:\\
1. **Scope**: Where a modifier or constraint applies\\
~~~(e.g., global vs filtered scope)\\
~~~- EXAMPLE: What activities does each gym offer?\\
~~~Interpretations:\\
~~~~~~- **1. Show classes that are common to all gyms.**\\
~~~~~~~~~~- SELECT Classes.Name FROM ...\\
~~~~~~~~~~- GROUP BY HAVING\\
~~~~~~~~~~- COUNT (DISTINCT Gyms\_Classes.GymID) =\\
~~~~~~~~~~~~(SELECT COUNT(*) FROM Gyms)\\
~~~~~~- **2. For each gym, show me the classes there.**\\
~~~~~~~~~~- SELECT Gyms.Name, Classes.Name FROM ...\\
2. **Attachment**: Which part of the sentence a modifier attaches to\\
~~~(e.g., prepositional phrase attachment)\\
~~~- EXAMPLE: Show the writers and editors on a work-for-hire.\\
~~~Interpretations:\\
~~~~~~- **1. Which writers are contracted for a work-for-hire\\
~~~~~~~~~~and which editors have the same contract?**\\
~~~~~~~~~~- SELECT Name FROM Screenwriters\\
~~~~~~~~~~- WHERE Contract = 'Work-for-Hire'\\
~~~~~~~~~~- UNION SELECT Name FROM Editors\\
~~~~~~~~~~- WHERE Contract = 'Work-for-Hire'\\
~~~~~~- **2. Provide a list of all writers and a list of\\
~~~~~~~~~~work-for-hire editors.**\\
~~~~~~~~~~- SELECT Name FROM Screenwriters\\
~~~~~~~~~~- UNION SELECT Name FROM Editors\\
~~~~~~~~~~- WHERE Contract = 'Work-for-Hire'\\
3. **Entity VAGUENESS**: Which entity or concept is being referred to\\
~~~(e.g., entity resolution, referential ambiguity)\\
~~~- EXAMPLE: Who issued CD Special?\\
~~~Interpretations:\\
~~~~~~- **1. Which bank issued CD Special?**\\
~~~~~~~~~~- SELECT Bank.Name FROM ...\\
~~~~~~~~~~- WHERE Products Name = 'CD Special'\\
~~~~~~- **2. Which branch issued CD Special?**\\
~~~~~~~~~~- SELECT Branch.Name FROM ...\\
~~~~~~~~~~- WHERE Products.Name = 'CD Special'\\
~~~~~~- **3. Find the bank and branch that issued CD Special.**\\
~~~~~~~~~~- SELECT Bank.Name, Branch.Name FROM ...\\
~~~~~~~~~~- WHERE Products.Name = 'CD Special'\\
\end{tcolorbox}
\caption{Ambiguity prompt-based filter (Part 1).}
\label{fig:prompt_filter_p1}
\end{figure*}

\begin{figure*}[t]
\begin{tcolorbox}[
    colback=gray!5!white,
    colframe=gray!75!black,
    title=Ambiguity Prompt-Based Filter (Part 2),
    fontupper=\small\ttfamily,
    width=\textwidth
]
4. **Broadness**: Sentence is vague and could be interpreted\\
~~~in multiple ways.\\
~~~- EXAMPLE: Show me some stats on the best player\\
~~~Interpretations:\\
~~~~~~1. Best player = most runs. stats = batting average, etc.\\
~~~~~~2. Best player = highest draft pick. stats = position, etc.\\
~~~~~~3. Best player = most wins. stats = wins, losses, etc.\\
~~~~~~many more\\
</ambiguity\_types>\\
\\
Here are a few more examples of ambiguity:\\
NL: Can you give me a breakdown of Terry Labonte's teams,\\
~~~average start and finish positions, and the most top 5\\
~~~finishes he had with each team between 1990 and 1999?\\
Ambiguity explanation: not clear if "between 1990-1999" applies\\
~~~to all conditions or just the last one\\
NL: What are the titles and directors of episodes that premiered\\
~~~in February and were written by writers who have worked on\\
~~~multiple episodes?\\
Ambiguity explanation: could be ones that BOTH premiered in\\
~~~February and were written by multi-episode writers, or EITHER\\
~~~condition\\
NL: Which NHL teams have the most Canadian players drafted from\\
~~~the OHL or WHL in the 1994 Entry Draft, and what positions\\
~~~do they play?\\
Ambiguity explanation: could mean both OHL 1994 or WHL 1994,\\
~~~or OHL (in general) vs WHL 1994\\
\\
Here are a few more examples of NOT valid ambiguity types:\\
NL: What is the total number of goals scored by George Utley\\
~~~during his time at Barnsley Football Club?\\
Ambiguity explanation: No valid alternative way to interpret.\\
NL: Which players have the best singles win-loss ratio in the\\
~~~Lithuania Davis Cup team?\\
Ambiguity explanation: No valid alternative way to interpret.\\
NL: Which year between 2005 and 2011 had the highest BBC Two\\
~~~viewership for Comic Relief?\\
Ambiguity explanation: No valid alternative way to interpret.\\
~~~The phrase "that year" makes the reference unambiguous.\\
\\
**REJECT** the following as NOT valid ambiguity types:\\
- Pattern matching variations (LIKE '\%X\%' vs LIKE 'X\%')\\
- Conditional priority differences\\
- Implementation style differences (subquery vs JOIN)\\
- Grouping semantics differences (unless true Scope/Entity)\\
- Table/schema confusion\\
- Minor formatting or syntax variations\\
\\
<scoring\_criteria>\\
- **1.0 (Valid):** SQL represents a valid semantic interpretation\\
~~~due to Scope, Attachment, Entity, or Broadness ambiguity.\\
- **0.0 (Invalid):** SQL contradicts the NL, differs only in\\
~~~implementation style, or is not a valid ambiguity type.\\
\\
**Key requirement**: Ambiguity must be genuine and semantic.\\
Implementation differences should be scored 0.0.\\
</scoring\_criteria>\\
\\
Output as JSON:\\
\{\\
~~~~"score": $<$float$>$\\
\}
\end{tcolorbox}
\caption{Ambiguity prompt-based filter (Part 2).}
\label{fig:prompt_filter_p2}
\end{figure*}

\begin{figure*}[t]
\begin{tcolorbox}[
    colback=gray!5!white,
    colframe=gray!75!black,
    title=Multi-Interpretation SQL Generation,
    fontupper=\small\ttfamily,
    width=\textwidth
]
You are an expert SQL developer.\\
\\
<task>\\
The provided NL question is ambiguous. The first SQL query\\
below provides ONE interpretation. Your job is to identify the\\
type of ambiguity and then generate a DIFFERENT SQL query that\\
provides a SECOND valid interpretation.\\
\\
<ambiguity\_types>\\
Valid ambiguity types:\\
1. Scope: Where a modifier or constraint applies\\
2. Attachment: Which part of the sentence a modifier attaches to\\
3. Entity Vagueness: Which entity or concept is being referred to\\
4. Broadness: Sentence is vague and could be interpreted in\\
~~~multiple ways\\
</ambiguity\_types>\\
</task>\\
\\
<inputs>\\
<nl\_question>\\
\{nl\_question\}\\
</nl\_question>\\
\\
<first\_interpretation\_sql>\\
\{first\_sql\_query\}\\
</first\_interpretation\_sql>\\
\\
<database\_schema>\\
\{table\}\\
</database\_schema>\\
</inputs>\\
\\
<instructions>\\
1. Analyze the ambiguity (Scope, Attachment, Entity Vagueness,\\
~~~or Broadness) and state which type is present or none\\
2. Construct a new SQL query that interprets the question\\
~~~differently than the first query (or "No ambiguity found"\\
~~~if none of the types are present).\\
3. Adhere to strict formatting: Use SQLite syntax, enclose\\
~~~columns in backticks, end with semicolon.\\
</instructions>\\
\\
Output as JSON:\\
\{\\
~~~~"reasoning": "Identify the ambiguity and how this new query\\
~~~~~~~~interprets it differently (Max 1 sentence).",\\
~~~~"sql\_query": "The new SQLite query"\\
\}
\end{tcolorbox}
\caption{Multi-interpretation SQL generation for ambiguous queries.}
\label{fig:prompt_ambiguity}
\end{figure*}

\begin{figure*}[t]
\begin{tcolorbox}[
    colback=gray!5!white,
    colframe=gray!75!black,
    title=Final Prompt-Based Quality Filter,
    fontupper=\small\ttfamily,
    width=\textwidth
]
You are an expert in SQL interpretation.\\
Your job is to determine if:\\
1. The SQL query is a valid interpretation of the NL question\\
2. The SQL query makes sense and is not unnecessarily redundant\\
3. The NL makes sense and matches something an actual human would ask\\
\\
EXAMPLES:\\
Fail example:\\
~~~~NL: Can you list the spacious classrooms in rooms 143, 547,\\
~~~~or 700, starting with the ones that can hold the most students?\\
~~~~SQL: SELECT `building`, `room number` FROM `classroom`\\
~~~~WHERE `capacity` > 40 AND `room number` IN ('143', '547', '700')\\
~~~~ORDER BY `capacity` DESC LIMIT 5;\\
~~~~This should be REJECTED (score 0.0) because the NL is weird.\\
~~~~The "spacious" meaning capacity > 40 is not easily inferable\\
~~~~and overall the NL just sounds confusing and wordy.\\
~~~~NOTE: Sometimes implications are ok. For example, if it said\\
~~~~"selecting only rooms with large capacities", that would\\
~~~~reasonably connect to a filter on capacity > X. But just\\
~~~~adding one adjective to a question is not enough to be\\
~~~~considered valid.\\
\\
Pass example:\\
~~~~NL: Can you give me a detailed report on the average height,\\
~~~~average weight, and number of perpetrators for each hometown,\\
~~~~sorted in descending order by the number of perpetrators, so\\
~~~~I can identify key trends and patterns?\\
~~~~SQL: SELECT `home town`, AVG(`height`) AS `average\_height`,\\
~~~~AVG(`weight`) AS `average\_weight`, COUNT(*) AS\\
~~~~`number\_of\_perpetrators` FROM `people` GROUP BY `home town`\\
~~~~ORDER BY `number\_of\_perpetrators` DESC;\\
~~~~This should be ACCEPTED (score 1.0) because the NL is clear\\
~~~~and the SQL makes sense. The NL implies that we are looking\\
~~~~for the average height, average weight, and number of\\
~~~~perpetrators for each hometown, sorted in descending order\\
~~~~by the number of perpetrators. The SQL does this and runs\\
~~~~successfully on the database schema.\\
\\
If all of the above are true, then the score should be 1.0,\\
otherwise it should be 0.0.\\
\\
NL: \{nl\}\\
SQL: \{sql\}\\
Database schema: \{database\_schema\}\\
\\
Output as JSON:\\
\{\\
~~~~"score": <float>,\\
~~~~"reasoning": "Why the score is what it is (max 2 sentences)"\\
\}
\end{tcolorbox}
\caption{Final prompt-based quality filter verifying NL-SQL alignment, SQL quality, and NL naturalness.}
\label{fig:prompt_final_filter}
\end{figure*}

\end{document}